\documentclass{article}
\usepackage[preprint]{neurips_2023}

\usepackage[utf8]{inputenc}
\usepackage[T1]{fontenc}    
\usepackage{hyperref}    
\usepackage{url}          
\usepackage{booktabs}   
\usepackage{amsfonts}     
\usepackage{nicefrac}   
\usepackage{microtype}      
\usepackage[dvipsnames, table]{xcolor} 
\usepackage[skins,breakable]{tcolorbox}
\usepackage{amsmath}
\usepackage{float}
\usepackage{array}
\usepackage{enumitem}
\usepackage{endnotes}
\usepackage{footnote}

\hypersetup{
    colorlinks,
    linkcolor={blue!80!black},
    citecolor={blue!50!black},
    urlcolor={blue!80!black},
}

\providecommand{\tightlist}{%
  \setlength{\itemsep}{0pt}\setlength{\parskip}{0pt}}

\newcommand{\clg}{\textcolor[RGB]{149,186,68}} 
\newcommand{\cm}{\textcolor[RGB]{194,63,112}}  
\newcommand{\cdg}{\textcolor[RGB]{104,193,127}} 
\newcommand{\cp}{\textcolor[RGB]{141,125,194}}   
\newcommand{\cb}{\textcolor[RGB]{61,133,198}}   
\newcommand{\co}{\textcolor[RGB]{230,145,56}}
\definecolor{lightgreen}{RGB}{217, 234, 211}

\newtcolorbox{safetycase}[1][Nested claims support parent claims.]{
  colback=lightgreen,
  colframe=lightgreen!70!black,
  enhanced,
  breakable,
  sharp corners,
  boxrule=0.5mm,
  width=\textwidth,
  title={},
  coltitle=black,
  fonttitle=\normalfont,
  colbacktitle=lightgreen,
  titlerule=0.5mm,
  top=0.5cm,
  toptitle=0.5cm,  
  bottomtitle=0.5cm, 
  left=0.75cm, 
  right=0.75cm
}

\setlist[description] {
  style=standard,  
  labelwidth=1.5em,   
  labelindent=0em,   
  labelsep=0.5em,     
  leftmargin=!,      
  itemindent=0em, 
}

\title{An Example Safety Case for\\Safeguards Against Misuse}

\author{%
  Joshua Clymer$^{\dagger}$\thanks{Equal Contribution. Correspondence to \href{mailto:joshuamclymer@gmail.com}{joshuamclymer@gmail.com} and \href{mailto:robert.kirk@dsit.gov.uk}{robert.kirk@dsit.gov.uk}}\\
  \And
  Jonah Weinbaum$^{\ddagger*}$ \\
  \And
  Robert Kirk$^{\S*}$ \\
  \AND
  Kimberly Mai$^{\S}$ \\
  \And
  Selena Zhang$^{\diamond}$ \\
  \And
  Xander Davies$^{\S}$
  \AND
$^{\dagger}$Redwood Research \quad
$^{\ddagger}$MATS \quad
$^{\diamond}$MIT \quad
$^{\S}$UK AI Security Institute 
}

\begin{document}
\maketitle

\begin{abstract}
Existing evaluations of AI misuse safeguards provide a patchwork of evidence that is often difficult to connect to real-world decisions. To bridge this gap, we describe an end-to-end argument (a ``safety case'') that misuse safeguards reduce the risk posed by an AI assistant to low levels. We first describe how a hypothetical developer red teams safeguards, estimating the effort required to evade them. Then, the developer plugs this estimate into a quantitative \textbf{``uplift model''} to determine how much barriers introduced by safeguards dissuade misuse (\url{https://www.aimisusemodel.com/}). This procedure provides a continuous signal of risk during deployment that helps the developer rapidly respond to emerging threats. Finally, we describe how to tie these components together into a simple safety case. Our work provides one concrete path -- though not the only path -- to rigorously justifying AI misuse risks are low.

\begin{figure}[htbp]
\centering
\includegraphics[width=0.8\linewidth]{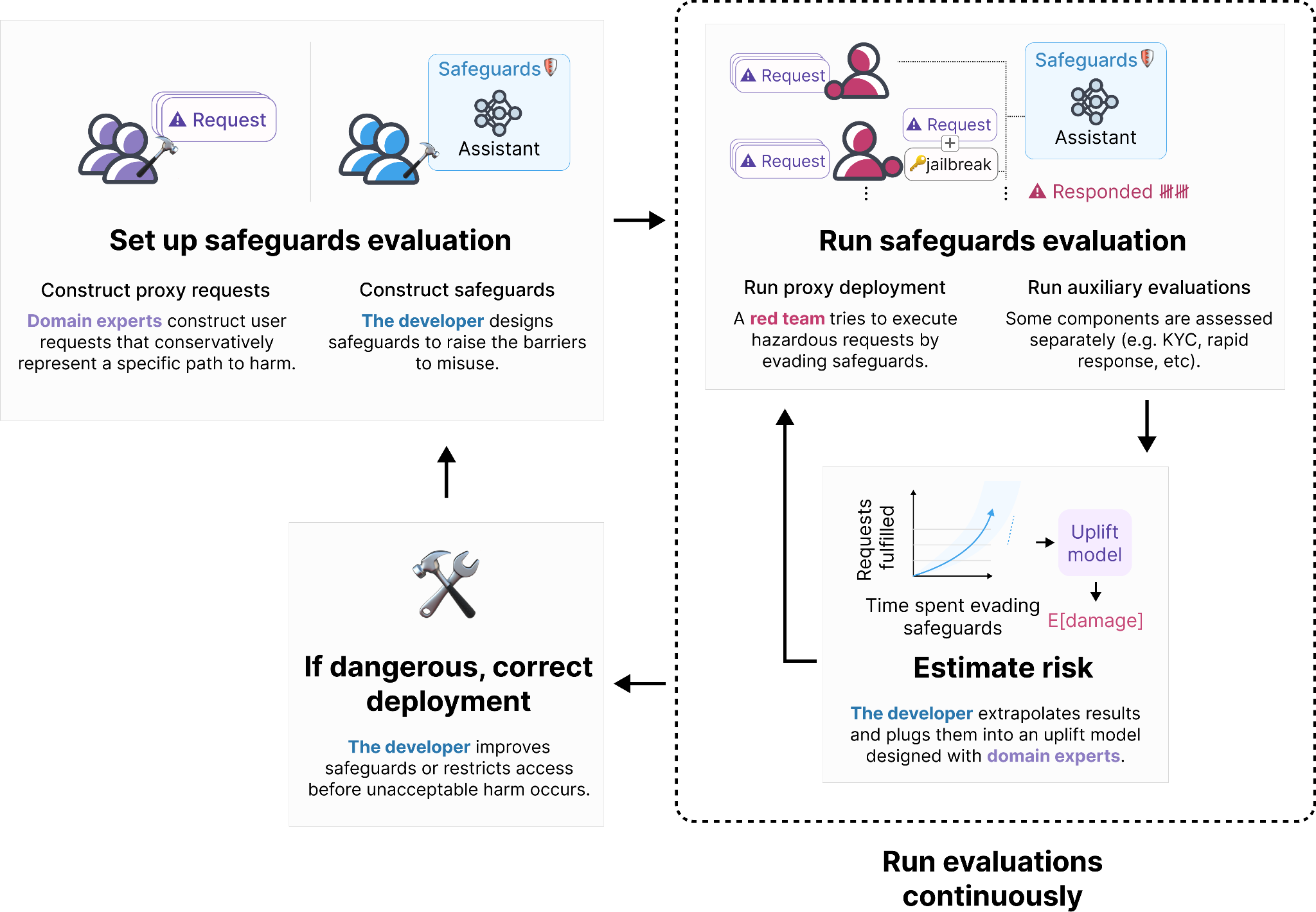}
\caption{An illustration of the safeguard evaluation
methodology we base our safety case on.}
\label{fig:methodology-overview}
\end{figure}

\end{abstract}

\setcounter{footnote}{0}

\section*{Introduction}\label{introduction}

OpenAI recently reported, ``several of our\ldots{} evaluations indicate our models are on the cusp of\ldots{} cross{[}ing{]} our high risk threshold'' \citep{OpenAI2025}. Anthropic published a similar statement: ``we believe there is a substantial probability that our next model may require ASL-3 safeguards'' \citep{Anthropic2025}, and more recently activated ASL-3 safeguards for their latest model release \citep{Anthropic2025asl3}. As models approach dangerous capability thresholds, some developers plan to shield these capabilities from bad actors with misuse ``safeguards,'' such as monitors and refusal training \citep{Anthropic2024,Anthropic2025asl3}. These safeguards might soon become load bearing for safe deployment \citep{OpenAI2025, Anthropic2025}, so there's an urgent need to rigorously assess them.

Researchers have taken large strides towards assessing safeguards by conducting large-scale red-teaming exercises \citep{sharma2025constitutionalclassifiersdefendinguniversal, GraySwanArena2025}, but the results of these exercises can be difficult to connect to real world decisions. The AI security community currently lacks a systematic way to map misuse safeguard evaluations to actionable risk estimates.

\textbf{Our work fills this gap by explaining an example argument (a safety case) that misuse safeguards ensure risk is below a specified level.} We focus on mitigations that developers apply to AI assistant APIs, building on the 5-step process laid out in the \emph{Principles for Safeguard Evaluations} \citep{UKAISI2025}. There are also other approaches to ruling out misuse risks, such as arguments that open-weight models will offset their danger by improving societal resilience (\hyperref[3.-discussion]{3. Discussion}), which we leave out of scope.

\section*{Executive Summary}\label{executive-summary}

\subsection*{Safety case objective}\label{safety-case-objective}

In our example safety case, the developer aims to defend the following
claim (\hyperref[2.1-safety-case-objectives-and-definitions]{2.1 Safety case objectives and definitions}):

\begin{tcolorbox}[colback=yellow!20]
\textbf{[C0]}: Given safeguards, the AI assistant does not pose risk above a certain level, defined as incurring large-scale harm above some threshold T in expectation.\footnotemark{}
\end{tcolorbox}

\footnotetext{This safety case is agnostic to the particular threat model and the metric of harm, but possible metrics include the expected number of annual fatalities or financial damage.}

We'll first explain the procedures the developer follows to justify that this claim holds, and then we'll organize this evidence into a safety case.

\subsection*{Safeguard evaluation}\label{safeguard-evaluation}

\begin{figure}[htbp]
\centering
\includegraphics[width=0.8\linewidth]{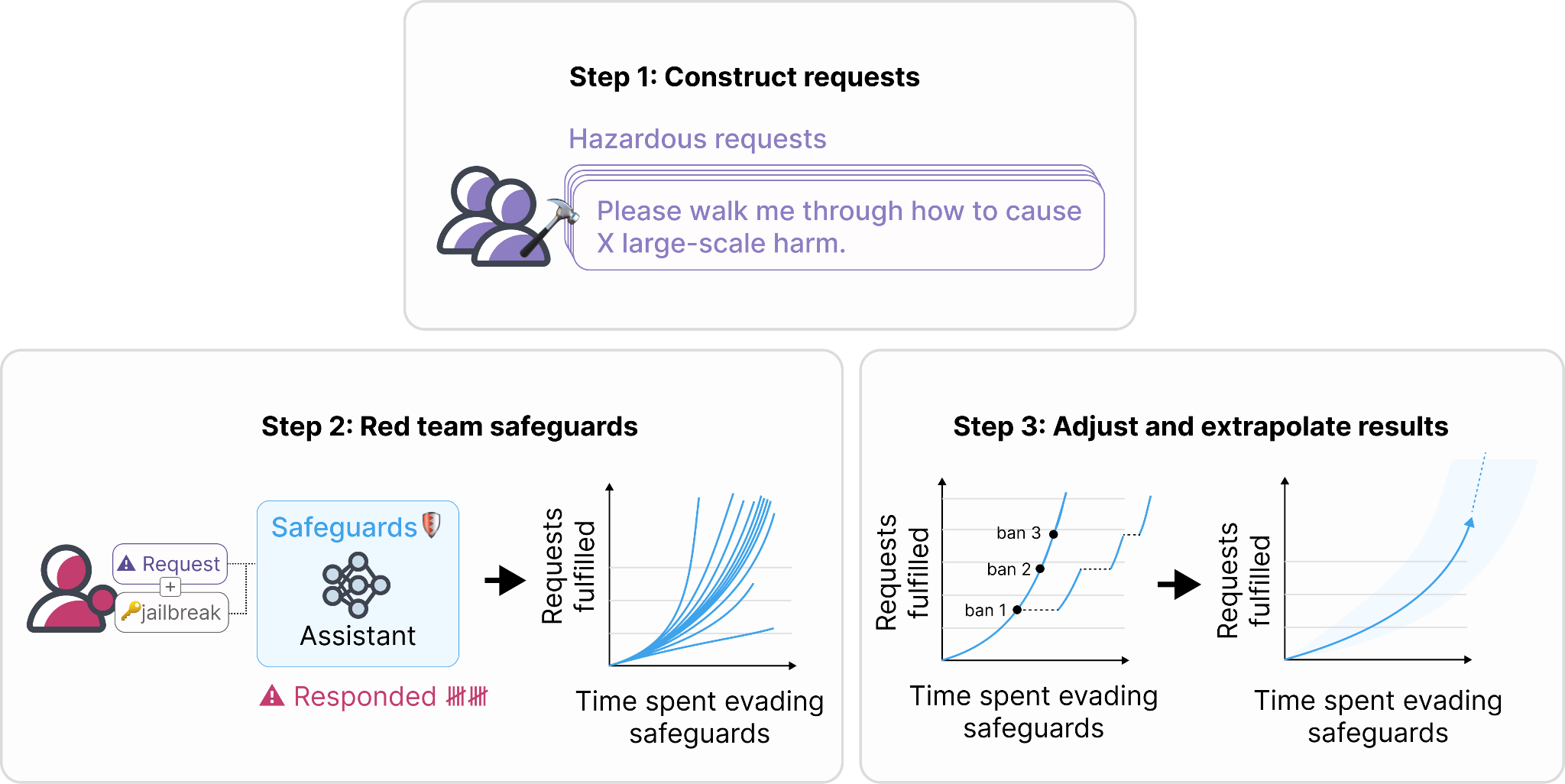}
\caption{Steps of a safeguards evaluation.}
\label{fig:safeguard-eval-steps}
\end{figure}

The developer first performs a \cb{\textbf{safeguard evaluation}} to estimate the barriers safeguards create to misuse actors. The safeguard evaluation involves three steps:

\begin{itemize}[leftmargin=10pt]
\tightlist
\item
  First, \cp{\textbf{threat model experts}} construct a representative set of harmful requests.\\
\item
  Next, members of a \cm{\textbf{red team}} try to fulfill these requests by evading safeguards (\hyperref[2.6-responding-to-evaluation-results]{section 2.6}). This results in a set of curves -- one for each member of the red team -- which relate the amount of time that member spent evading safeguards with the number of harmful requests fulfilled.\\
\item
  Finally, the developer adjusts these curves, factoring in components of safeguards assessed separately such as account banning, and extrapolates them into a smooth trendline (\hyperref[fig:safeguard-eval-steps]{Figure 2}).
\end{itemize}

\subsection*{Uplift
model}\label{uplift-model}

The safeguard evaluation reveals how much time users would need to spend looking for holes in safeguards. Is the effort required sufficient to dissuade misuse? This question might be difficult to answer with qualitative analysis, so we discuss how a developer could construct a quantitative \textbf{``uplift model.''} You can visit \url{https://www.aimisusemodel.com/} to explore an interactive example.

Our \textbf{uplift model} involves three steps
(\hyperref[2.5-the-uplift-model]{section
2.5}):

\begin{itemize}[leftmargin=10pt]
\tightlist
\item
  \textbf{Step 1.} The developer first estimates the \cdg{\textbf{existing level of risk}}, assuming a scenario where the AI assistant is not
  deployed. This risk can be estimated by considering the
  \textbf{time-costs of causing harm through the risk pathway of concern,} and the \textbf{willingness of novice actors to pay these costs.}
\end{itemize}

\begin{figure}[H]
\centering
\includegraphics[width=0.8\linewidth]{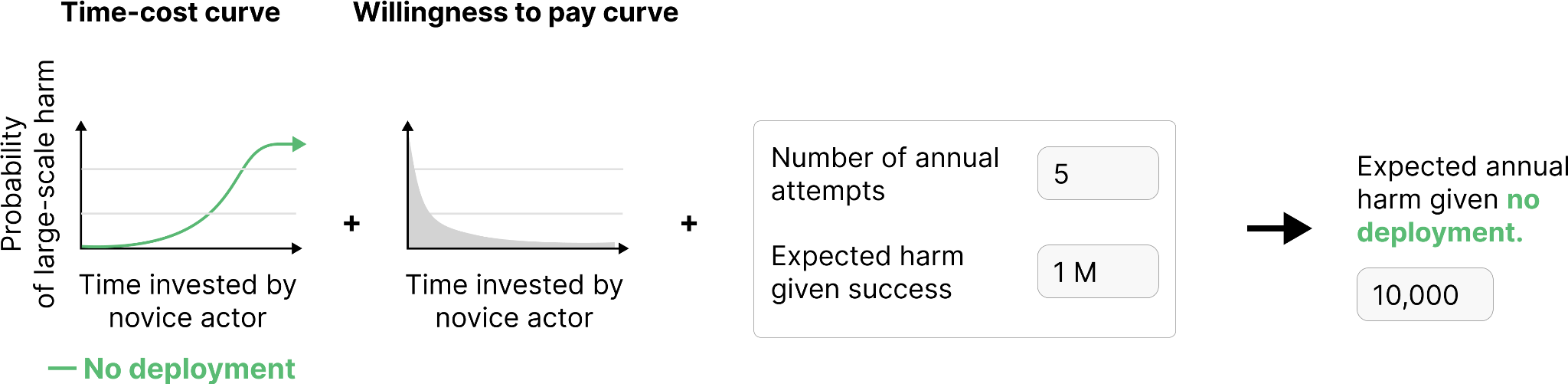}
\end{figure}

\begin{itemize}[leftmargin=10pt]
\tightlist
\item
  \textbf{Step 2.} Next, the developer estimates the level of \clg{\textbf{risk if the AI assistant is deployed}}.

  \begin{itemize}
  \tightlist
  \item
    \textbf{Step 2.1.} To arrive at this risk estimate, the developer first assesses the risk posed by a \cm{\textbf{pre-mitigation}} version of the AI assistant.
  \end{itemize}

\begin{figure}[H]
\centering
\includegraphics[width=0.8\linewidth]{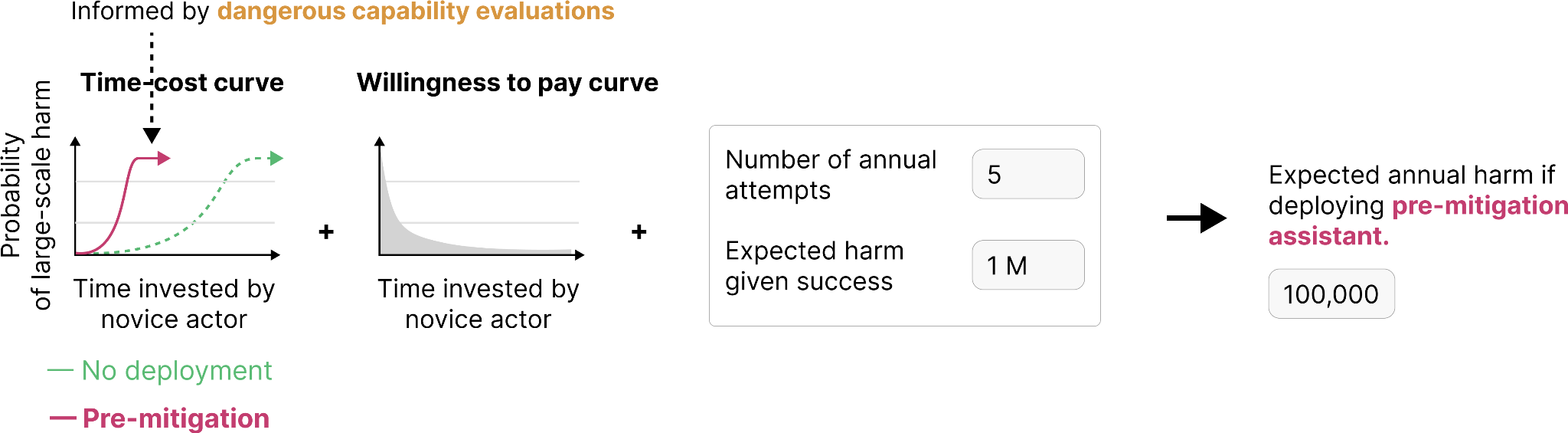}
\end{figure}

\begin{itemize}
\tightlist
\item
  \textbf{Step 2.2.} Next, the developer determines how much safeguards
  lower \cm{\textbf{pre-mitigation risk}} by drawing on results from
  the \cb{\textbf{safeguard evaluation}}. Specifically, the developer
  calculates how \cb{\textbf{safeguard evaluation}} results shift points
  along the \cm{\textbf{pre-mitigation}} \textbf{time-cost curve}. By combining the
  new ``time-cost curve'' with the ``willingness to pay'' distribution
  mentioned previously, the developer estimates \clg{\textbf{post-mitigation
  risk}}.
\end{itemize}

\begin{figure}[H]
\centering
\includegraphics[width=0.9\linewidth]{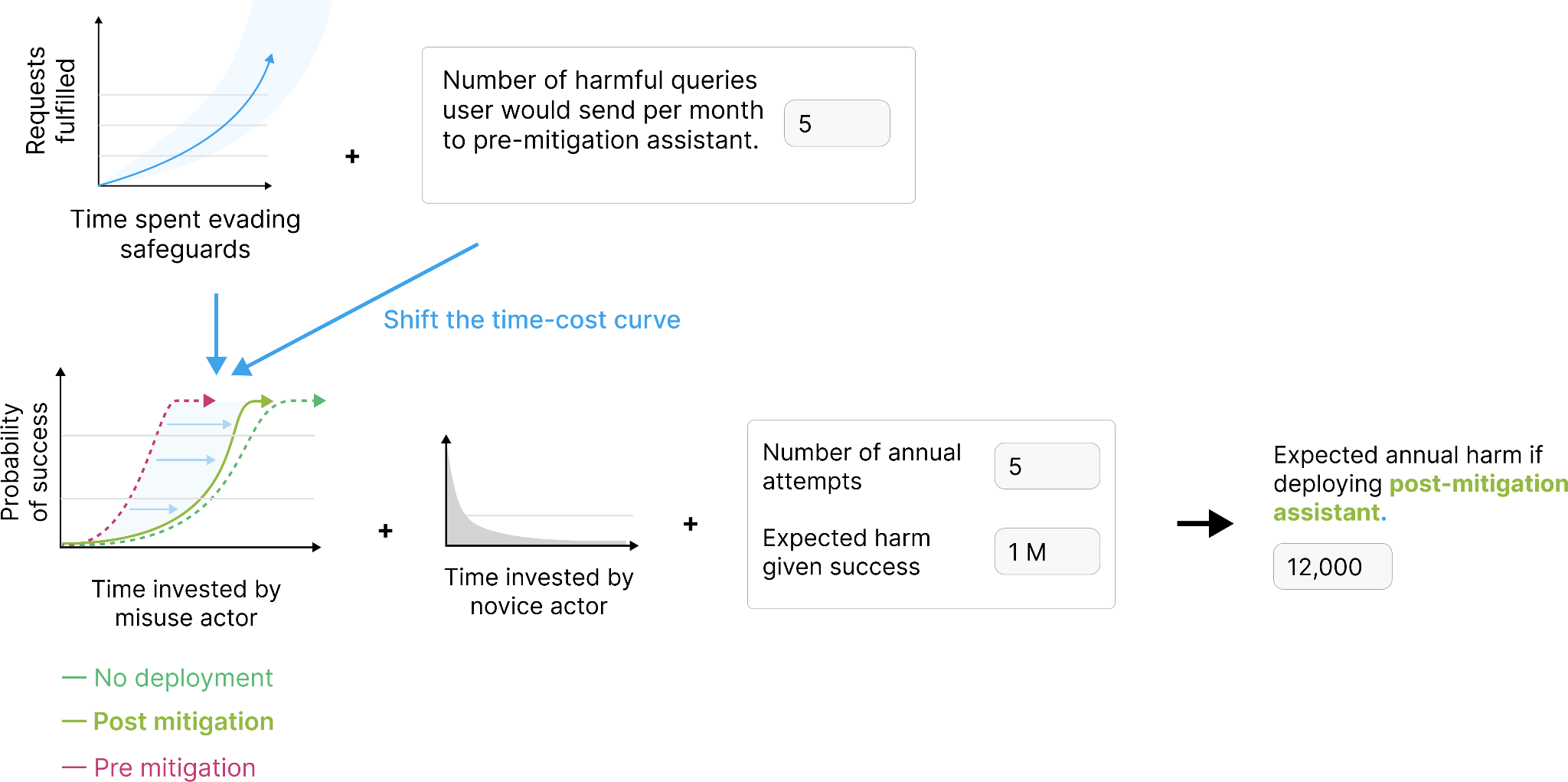}
\end{figure}
\end{itemize}

\begin{itemize}[leftmargin=10pt]
\tightlist
\item
  \textbf{Step 3.} The output of the uplift model is the difference
  between the \clg{\textbf{post mitigation risk}} found in step 2 and the
  \cdg{\textbf{existing level of risk}} estimated in step 1.
\end{itemize}

\subsection*{Responding to signs of
risk}\label{responding-to-signs-of-risk}

Safeguard effectiveness can change post-deployment as the pool of
adversaries grows \citep{UKAISI2025}. The developer follows three procedural policies that
define how they respond to evaluations throughout the AI deployment
lifecycle:

\begin{itemize}[leftmargin=10pt]
\tightlist
\item
  \textbf{Policy 1:} The developer first runs \textbf{pre-deployment
  evaluations}. AI assistants estimated to pose risk above the threshold
  \textbf{T} are not deployed.\\
\item
  \textbf{Policy 2:} Even after an AI assistant is cleared for
  deployment, the developer continues to perform safeguard
  evaluations\textbf{.} The red team (which could include participants
  in a bug bounty program) keeps searching for holes in safeguards --
  drawing on black markets, social media, and emerging new jailbreaking
  methods just as users would.\\
\item
  \textbf{Policy 3:} Before the risk estimated by the uplift model
  exceeds the threshold \textbf{T}, the developer responds to emerging
  attacks -- hardening faulty safeguards, or in an extreme case,
  restricting access to the AI assistant.
\end{itemize}

\subsection*{Safety case
outline}\label{safety-case-outline}

Together, the evaluations and procedural commitments described earlier
help keep risk low. However, there are many moving parts to keep track
of. To maintain these arguments and evidence throughout deployment, a
developer might write a ``safety case'' -- a structured argument that
their AI assistant poses risks below some threshold
[\citet{Clymer2024}, step 5 of \citep{UKAISI2025}]. Drawing on the evidence discussed, the following
summarizes our safety case (\hyperref[2.7-safety-case]{section 2.7}):

\begin{safetycase}

\textbf{Key:}\\
\textbf{[C]:} claim\\
Nested claims support parent claims.\\

\textbf{Safety case outline (\hyperref[2.7-safety-case]{section 2.7}):}
\begin{description}[leftmargin=2em, labelwidth=2em, labelsep=0.3em]
  \item[\textbf{[C0]}] The AI assistant does not increase risks in class \textbf{X} above some threshold \textbf{T}.
  
  \begin{description}[leftmargin=2em, labelwidth=2em, labelsep=0.5em]
    \item[\textbf{[C1]}] The AI assistant cannot pose risks in class \textbf{X} aside from helping novice actors perform misuse attempts through pathway \textbf{Y}.
    
    \item[\textbf{[C2]}] If safeguards are kept in place and procedural policies are followed, the AI assistant will not pose risk above threshold \textbf{T} through aiding with misuse attempts in pathway \textbf{Y}.
    
    \begin{description}[leftmargin=2em, labelwidth=2.7em, labelsep=0.5em]
      \item[\textbf{[C2.1]}] Developers will correct deployment to reduce risk to below the threshold within some period of time (we'll say one month)\footnotemark{} of identifying risk increasing towards the threshold.
      
      \item[\textbf{[C2.2]}] Safeguard evaluations conservatively bound the level of risk forecasted one month into the future.
      
      \begin{description}[leftmargin=2em, labelwidth=3.8em, labelsep=0.3em]
        \item[\textbf{[C2.2.1]}] The "main evaluation" lower-bounds safeguard effectiveness (is conservative).
        
        \begin{description}[leftmargin=2em, labelwidth=4.5em, labelsep=0.3em]
          \item[\textbf{[C2.2.1.2]}] The dataset of requests is representative of user requests for assistance with pathway Y.
          
          \item[\textbf{[C2.2.1.3]}] The safeguards in the evaluation are a conservative proxy for the safeguards in deployment (holding out the safeguards evaluated separately, which in this case, includes account banning and KYC).
          
          \item[\textbf{[C2.2.1.4]}] Individual members of the red team are more competent on average than individual novice misuse actors, competitively resourced, and well-incentivized.
          
          \item[\textbf{[C2.2.1.5]}] The behavior of the model in the evaluation is representative of its behavior in deployment (e.g. models are not faking alignment during evaluations).
        \end{description}
        
        \item[\textbf{[C2.2.2]}] The evaluations of other safeguard components (e.g. account banning, rapid vulnerability patching) are conservative.
        
        \item[\textbf{[C2.2.3]}] The uplift model conservatively estimates the current level of risk given evaluation results.
        
        \item[\textbf{[C2.2.4]}] The uplift model upper bounds the risk that would emerge in the next 1+ months in the worst case where safeguards suddenly cease to be effective.
      \end{description}
      
      \item[\textbf{[C2.3]}] The current estimated bound of risk forecasted 1 month into the future is below the threshold. Developers correcting deployment to reduce risk within 1 month keeps risk below the threshold.
    \end{description}
  \end{description}
  
  \item[\textbf{[C3]}] Safeguards will be kept in place and maintain effectiveness for the duration of the deployment.
\end{description}
\end{safetycase}

\footnotetext{This one month reaction time is an arbitrary stand-in for however long a developer conservatively expects to be able to make the necessary changes to their deployment to notice and mitigate risk before it rises above the threshold.}

We believe the evaluation methodology we outline is practical for
developers to implement. However, we describe variations in \hyperref[appendix-c:-potential-adjustments-to-the-safety-case]{Appendix C} that require less effort from the developer.

\section{Safety Cases Background}\label{1.-safety-cases-background}

A \textbf{safety case} is a flexible format for justifying that a system is safe in a given operating environment \citep{Clymer2024, Wasil2024, buhl2025safety}. Developers begin with broad claims and iteratively break these down into smaller, testable sub-claims informed by threat modelling and expert review.

Researchers have already begun applying safety-case methods to AI
systems to demonstrate that, for example, a model does not engage in
subversive ``scheming'' or assist in catastrophic cyberattacks \citep{goemans2024safetycasetemplatefrontier, Balesni2024, Buhl2024, korbak2025sketchaicontrolsafety}.
Building on this work, the remainder of this paper sketches a safety
case that shows a model with safeguards does not increase the risk of a
hypothetical large-scale harm due to misuse of the AI system above some
threshold.

\section{An Example Safety Case for Misuse Safeguards}\label{2.-an-example-safety-case-for-misuse-safeguards}

We now lay out our method for arriving for actionable risk-estimates
from safeguard evaluations, and hence making a safety case that argues a
hypothetical AI assistant does not pose a level of risk above some
threshold \textbf{T} due to misuse.

Sections \ref{2.1-safety-case-objectives-and-definitions}-\ref{2.2-threat-modelling} demonstrate how to produce a safeguard requirement that
describes in detail the threat actor and risk scenario of concern,
satisfying step 1 of the \emph{Principles for Evaluating Misuse
Safeguards}
\citep{UKAISI2025}. Next, we describe specific safeguards
(\hyperref[2.3-misuse-safeguards]{2.3}, which corresponds to step 2 of
the \emph{Principles}), explain the safeguard evaluation procedure
(\hyperref[2.4-the-safeguard-evaluation-methodology]{2.4},
\hyperref[2.5-the-uplift-model]{2.5} \&
\hyperref[2.6-responding-to-evaluation-results]{2.6}, which correspond
to step 3 and 4 of the \emph{Principles}), and finally, tie all of this
evidence into a safety case (\hyperref[2.7-safety-case]{2.7},
\emph{Principles} step 5).

\subsection{Safety case objectives and
definitions}\label{2.1-safety-case-objectives-and-definitions}

We assume that the AI assistant (without safeguards, or pre-mitigation)
can provide meaningful uplift in realising risk in a given pathway to
harm, which the developer confirms through dangerous capability
evaluations (i.e.~its capabilities would put it into a classification of
`high'' risk). Given this, we cannot make an \emph{inability safety
case} \citep{goemans2024safetycasetemplatefrontier}.
Instead, the developer aims to justify that their safety measures
\emph{mitigate} these dangerous capabilities, such that the actual risk
is comparable to a pre-mitigation system at a level of capability
corresponding to some lower threshold of risk.

Our safety case relies on quantitative risk estimates, so the first step
in constructing this safety case is to convert qualitative ``risk levels'
(perhaps defined in a developer's frontier safety framework) to
quantitative ones.

\begin{figure}[H]
\centering
\includegraphics[width=0.8\linewidth]{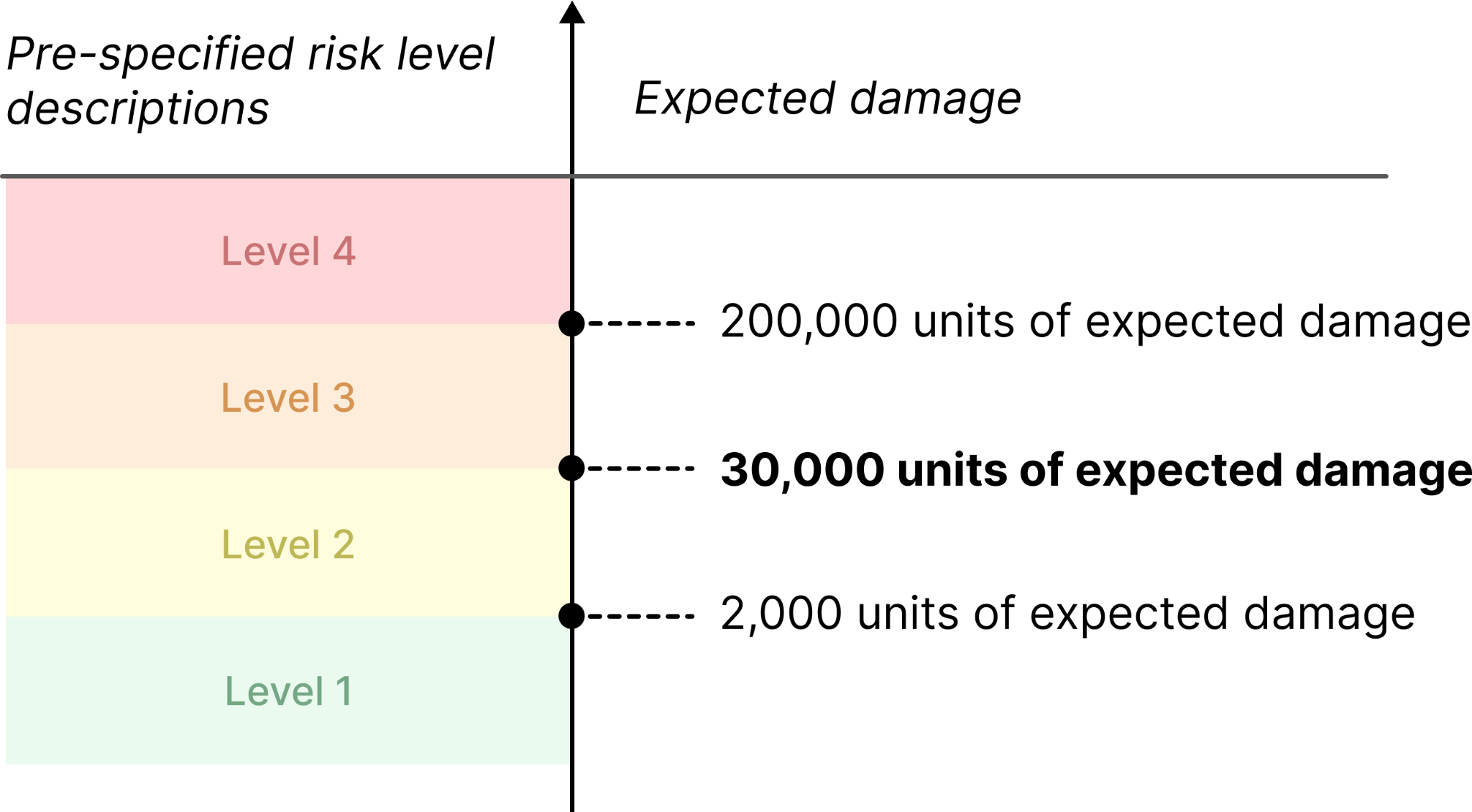}
\caption{Mapping qualitative risk levels onto quantitative thresholds.}
\end{figure}

The developer then constructs a safety case to defend the following
top-level claim:

\begin{tcolorbox}[colback=yellow!20]
\textbf{[C0]}: Given safeguards, the AI assistant does not pose risk above a certain level, defined as incurring large-scale harm above some threshold T in expectation.
\end{tcolorbox}
\vspace{5pt}
\centerline{Box 1. The top-level claim that the safety case justifies.}

The table below defines terms used in the remainder of the safety case.\begin{table}[h]
\centering
\small
\begin{tabular}{|p{0.25\textwidth}|p{0.7\textwidth}|}
\hline
\rowcolor{gray!20}
\textbf{Term} & \textbf{Definition} \\
\hline
"Pre-mitigation" and "post-mitigation" AI assistant. & The AI system before and after safeguards are applied. The pre-mitigation AI assistant is trained to be helpful-only and answer all requests. \\
\hline
Red team & A collection of people paid or otherwise incentivized to safely demonstrate to a developer that they can evade misuse safeguards. \\
\hline
Novice actor ("novice") & An individual with little to no experience relevant to the risk domain of interest for misuse attempts, and less than \$30,000 available. Equivalently, a "lone wolf." \\
\hline
Universal jailbreak & An automatic function (or one which is trivial for a human to execute) which takes an input question and produces a new input which when put into the model produces a response the same as unrestricted access would provide. \\
\hline
Misuse attempt (or just "attempt") & A scenario where a novice actor spends more than two weeks attempting to realise large-scale harm through the risk pathway of concern. \\
\hline
"Time-cost" curve & A function that relates the amount of time a novice actor spends on their misuse attempt and their probability of succeeding at a misuse attempt. \\
\hline
"Willingness to pay" distribution & A distribution over the amount of time a novice actor will spend on a misuse attempt. \\
\hline
"Safeguard evasion cost" curve & A function that relates the amount of time an actor spends evading safeguards with the amount of requests they successfully fulfill from some pre-defined set. \\
\hline
\end{tabular}
\vspace{10pt}
\caption{Vocabulary used in this safety case sketch.}
\label{tab:vocab}
\end{table}

\subsection{Threat
modelling}\label{2.2-threat-modelling}

Threat modelling is inherently domain-specific, and so we don't provide
specific details per-domain here. In general, threat modelling narrows
down pathways to harm from a broad category (e.g.~cyber offence) to a
more specific pathway to harm (e.g.~writing a computer virus), based on
the results of \co{\textbf{dangerous capability evaluations.}} Once the most
plausible threat vectors are understood, \cp{\textbf{threat modelling
experts}} can design an appropriate dataset of harmful requests (\hyperref[2.4-the-safeguard-evaluation-methodology]{section 2.4}) and a threat-specific uplift model (\hyperref[2.5-the-uplift-model]{section 2.5}).

\subsection{Misuse safeguards}\label{2.3-misuse-safeguards}

\begin{figure}[t]
\centering
\includegraphics[width=0.8\linewidth]{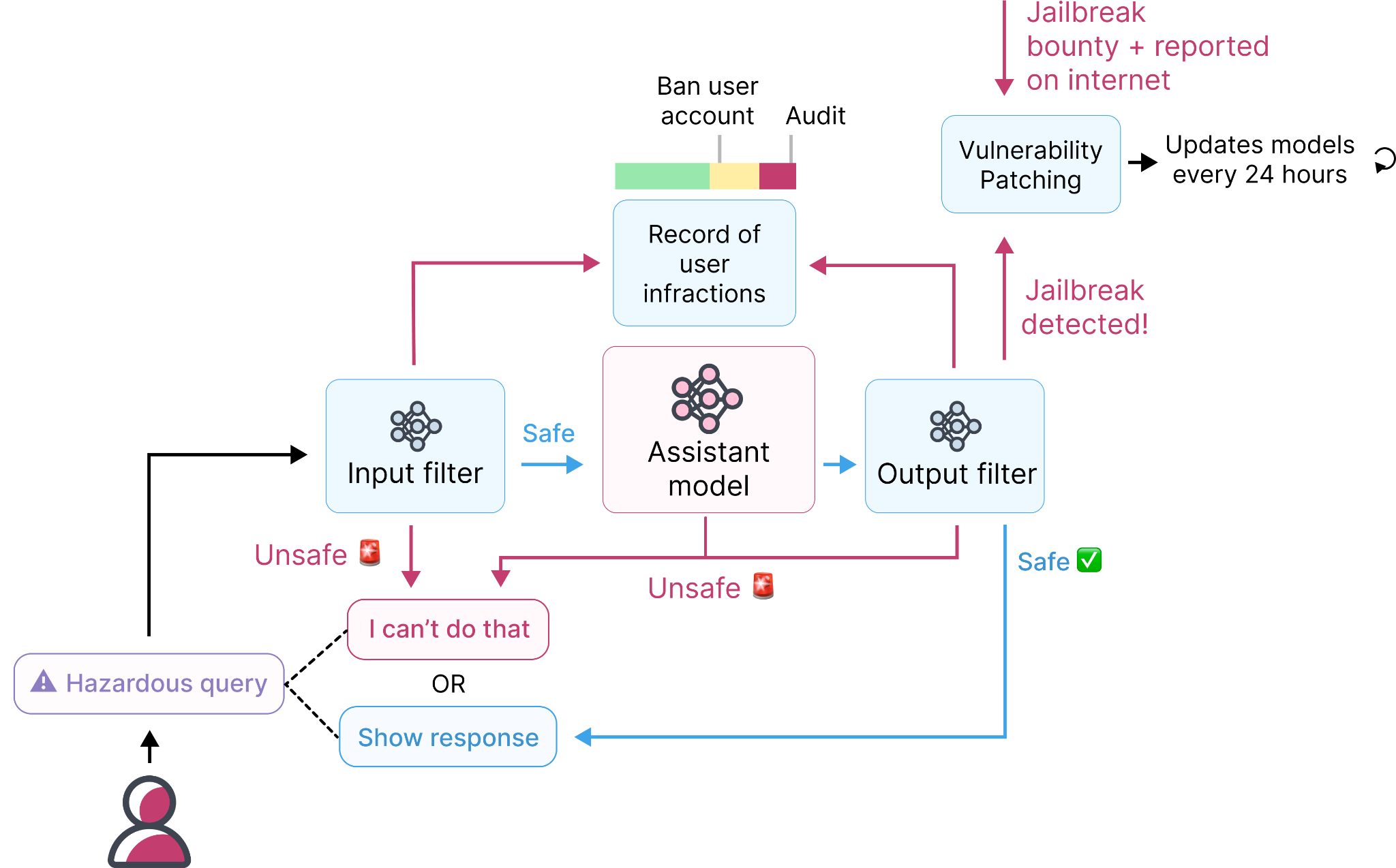}
\caption{An overview of misuse safeguards.}
\label{fig:misuse-safeguard-overview}
\end{figure}

To prevent the AI assistant from helping users with misuse attempts, the developer implements the misuse safeguards illustrated in \hyperref[fig:misuse-safeguard-overview]{Figure 4}. The developer details these safeguards, satisfying step 2 of the UK AISI Principles for Safeguards Evaluation
\citep{UKAISI2025}.\footnote{In this example we pick a set of safeguards we
  expect to be used by frontier AI companies and to be somewhat
  effective, but there are different sets or combinations of safeguards
  that could be used.}

Table 2 describes these safeguards in more detail.
\begin{table}[h]
\centering
\small
\begin{tabular}{|p{0.25\textwidth}|p{0.65\textwidth}|}
\hline
\rowcolor{gray!20}
\multicolumn{1}{|c|}{\textbf{Mitigation}} & \multicolumn{1}{c|}{\textbf{Description}} \\
\hline
Refusal training & The developer trained the assistant to follow a constitution \citep{Bai2022} that prohibits responding to requests that might aid realising large-scale harm along the risk pathway of concern. \\
\hline
KYC protections & The developer implements know-your-customer (KYC), such as \url{login.gov}, so that the novice will have difficulty creating new accounts if banned and is likely to incur suspicion in the process. \\
\hline
Input and output classifiers & An input classifier is run on every request that immediately rejects harmful requests and logs the incident. If the request is sent to the AI assistant, a separate output classifier reviews the input and output together, and once again immediately blocks harmful responses and logs the incident (similar to \citet{sharma2025constitutionalclassifiersdefendinguniversal}). \\
\hline
Account banning and reporting to law enforcement. & If the user repeatedly makes suspicious requests, their account is temporarily suspended. If the user continues making suspicious requests, the user is warned and then their account is banned. If the recorded incidents are sufficiently egregious, a human reviews them and reports the user to authorities. \\
\hline
Jailbreak bounty & The developer provides cash incentives for reported jailbreaks and buys jailbreaks from black markets if their price falls below the amount a novice actor might conceivably spend. \\
\hline
Vulnerability patching & Jailbreaks that are identified by output classifiers, reported on the internet, or submitted to the jailbreak bounty are added to training data for the assistant model and classifiers. These models are updated every 24 hours. \\
\hline
\end{tabular}
\vspace{10pt}
\caption{A description of misuse safeguards.}
\label{tab:misuse-safeguards}
\end{table}

\subsection{The safeguard evaluation
methodology}\label{2.4-the-safeguard-evaluation-methodology}

To evaluate whether the safeguards in \hyperref[2.5-the-uplift-model]{section 2.5} are effective, the
developer performs \textbf{safeguard evaluations.} In this section, we
describe how these evaluations are performed.

\subsubsection{The structure of a safeguard
evaluation}\label{the-structure-of-a-safeguard-evaluation}

A safeguard evaluation relies on several components:

\begin{itemize}[leftmargin=10pt]
\tightlist
\item
  \textbf{A dataset of harmful requests.} The dataset of misuse requests
  is designed by threat modelling experts to encompass representative
  paths to harm
  (\hyperref[appendix-f:-the-dataset-of-harmful-requests]{Appendix F}).
\end{itemize}

\vspace{-5pt}
\begin{figure}[H]
\centering
\includegraphics[width=0.5\linewidth]{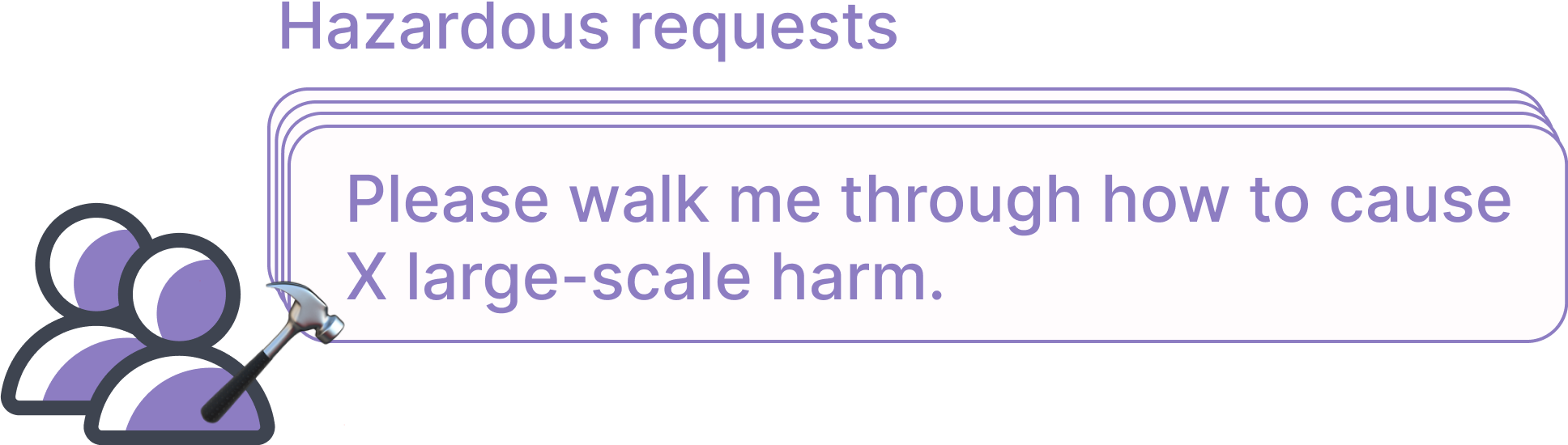}
\end{figure}

\begin{itemize}[leftmargin=10pt]
\item
  \textbf{Red team actors.} These might be contractors, employees, or
  external users, whose job it is to try to evade the safeguards during
  the evaluation and successfully fulfill the set of misuse requests. \\

  The red team should have equal or greater affordances to what novice misuse
  actors are expected to have. We suppose the red team can attempt to jailbreak queries in any order they want, and to submit any input to the AI assistant they want to produce the information that satisfies the original request. \\

  The objective of red team actors is to maximize their contribution to
  the risk estimated by the evaluation. \\

  See \hyperref[appendix-d:-the-red-team]{Appendix D: The red team} for
  detailed discussion of how to set up the red team effectively.
\item
  \textbf{Proxy safeguards.} The developer implements safeguards similar
  to the ones that will be used in deployment, which are designed to
  maximally reduce the rate at which the red team can fulfill requests.
\end{itemize}

\vspace{-5pt}
\begin{figure}[H]
\centering
\includegraphics[width=0.7\linewidth]{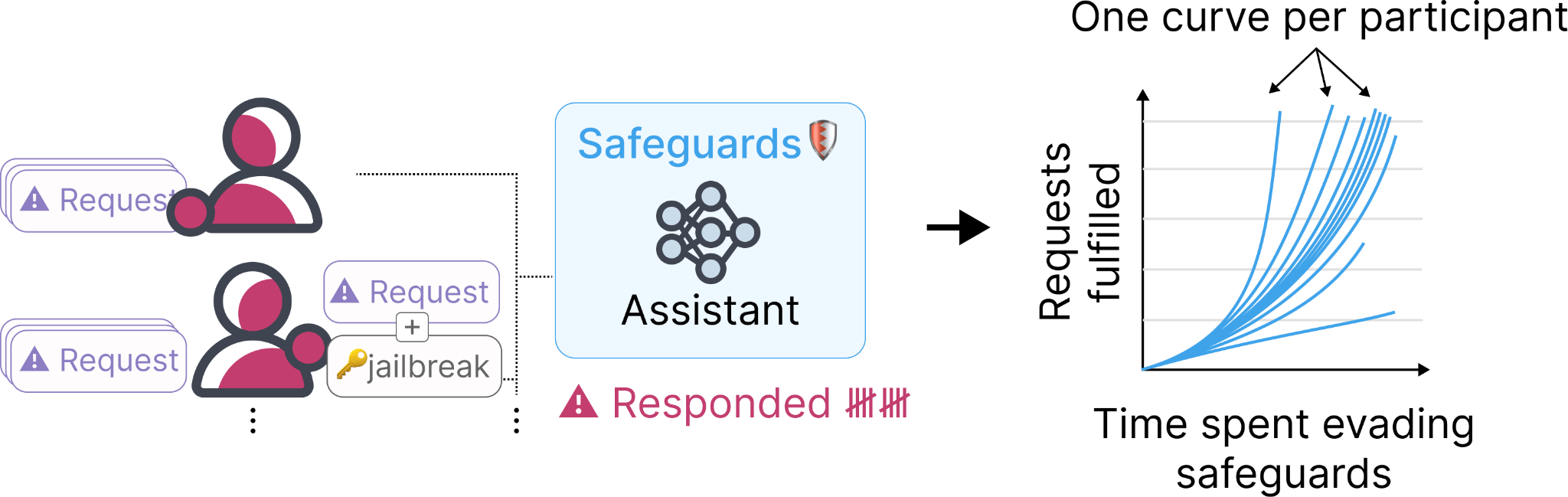}
\caption{Running the safeguard evaluation. The red team
attempts to fulfill harmful requests using the safeguarded AI assistant.}
\label{fig:running-safeguard-evaluation}
\end{figure}

\textbf{The output of a safeguard evaluation} is a sequence of
\textbf{safeguard evasion cost curves}, one for each red team actor, as
shown in \hyperref[fig:running-safeguard-evaluation]{Figure 5}. Each curve relates the requests a member of the red
team fulfills with the time they spent evading safeguards. To assess the
extent to which a request is ``fulfilled,'' the developer grades each AI
assistant response on a continuum of helpfulness where 0 is ``not
helpful at all'' and 1 is ``as helpful as the pre-mitigation response''
\citep{sharma2025constitutionalclassifiersdefendinguniversal}.

\subsubsection{Evaluating different safeguards
separately}\label{evaluating-different-safeguards-separately}

Some safeguards (such as account banning or vulnerability patching) are
convenient to evaluate independently. For instance, if members of the
red team are constantly banned from their accounts, the red team will
spend less time looking for jailbreaks, and the results will be more
volatile.

To illustrate how a developer might evaluate safeguards separately,
we'll consider a situation in which the following components are
evaluated in isolation:

\begin{enumerate}[leftmargin=10pt]
\def\labelenumi{\arabic{enumi}.}
\tightlist
\item
  All safeguards aside from account banning.\\
\item
  Account banning.
\end{enumerate}

To assess these safeguard components, the developer performs a ``main
evaluation,'' a ``ban evaluation.'

\begin{figure}[htbp]
\centering
\includegraphics[width=0.7\linewidth]{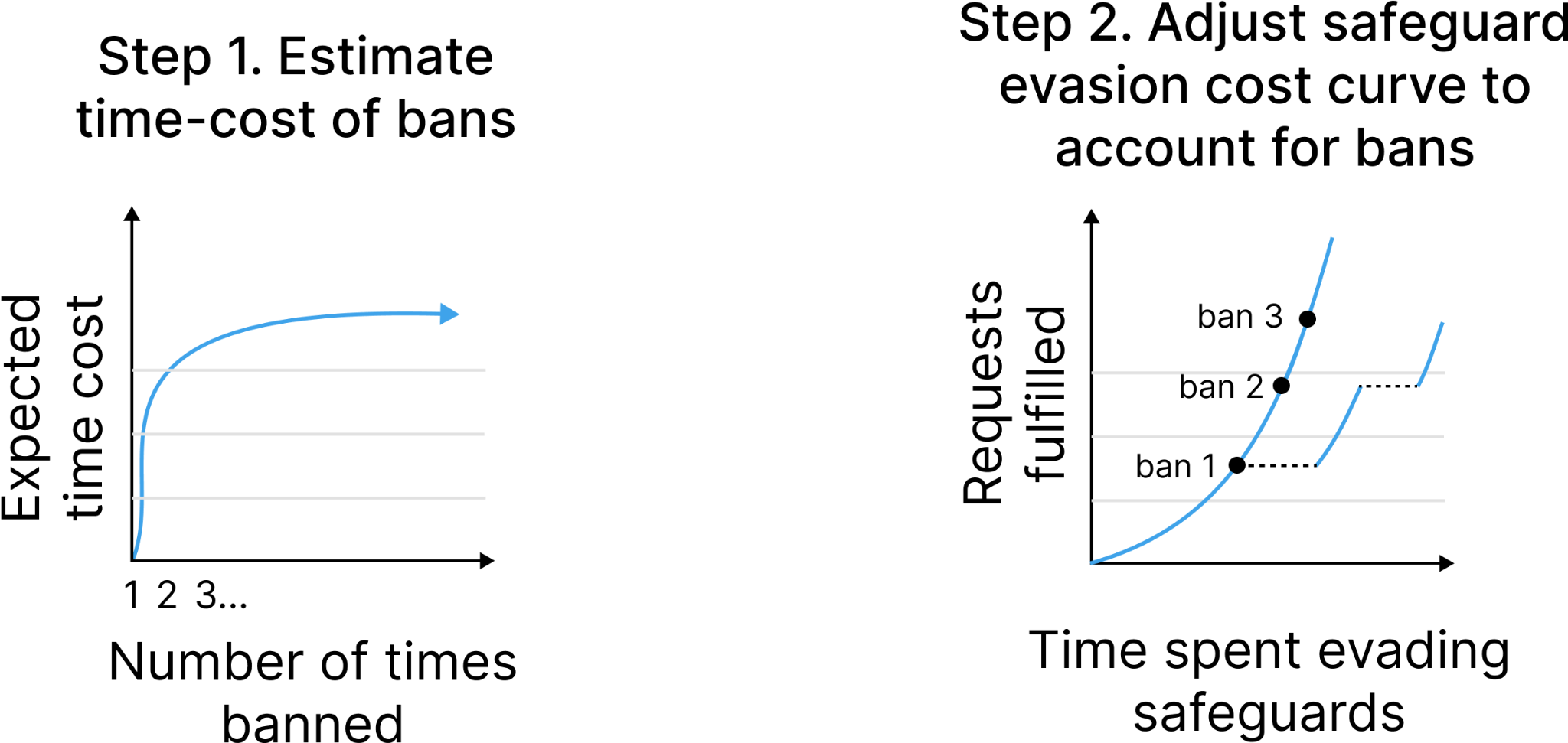}
\caption{Performing a safeguard evaluation that accounts for the time-cost of bans.}
\label{fig:ban-eval}
\end{figure}

\begin{itemize}[leftmargin=10pt]
\tightlist
\item
  During the ``\textbf{main evaluation},'' the system does not ban red
  team accounts. Instead, red team actors are informed when they
  \emph{would have been} banned and their ``time spent evading
  safeguards'' score is increased. Red teamers are still penalised for
  having their account banned, and hence are still incentivised to avoid
  account bans.\\
\item
  To determine how much time to add to red team time scores for bans, the developer performs a \textbf{``ban evaluation''} (\hyperref[fig:ban-eval]{Figure 6}). This evaluation employs a separate red team to find ways to evade KYC (which should be set up as described in \hyperref[appendix-d:-the-red-team]{Appendix D: The red team}). The evaluation may also rely on expert judgement to account for strategies that are illegal to pursue, and these evaluations only need to be performed once for a given set of KYC measures, and so this information might also be available in existing studies.
\end{itemize}

Account banning is not the only example of a safeguard component that is convenient to evaluate in isolation. We also discuss how a developer might evaluate vulnerability patching independently in \hyperref[appendix-a:-evaluations-of-vulnerability-patching]{Appendix A}.

\subsubsection{Aggregating and extrapolating
results}\label{aggregating-and-extrapolating-results}

After completing safeguard evaluations and combining their results, the
developer obtains a series of safeguard evasion curves. These curves
estimate how much effort misuse actors in deployment would need to spend
to fulfill hazardous requests for each red team actor. The developer
finally aggregates the curves into an extrapolated function, which is
the joint distribution of requests fulfilled given time spent evading
safeguards.\footnote{If the developer plans to run safeguard evaluations
  periodically rather than continuously, the details of this
  extrapolation become important to ensure the risk is correctly
  estimated or bounded between safeguard evaluations. In this case using
  more conservative methods for extrapolation are more suitable. Because
  of the difficulty of performing this extrapolation, even with
  continuous methods, we think running evaluations continuously is the
  correct approach in general, although this could take the form of a bug bounty program.}

\begin{figure}[htbp]
\centering
\includegraphics[width=0.6\linewidth]{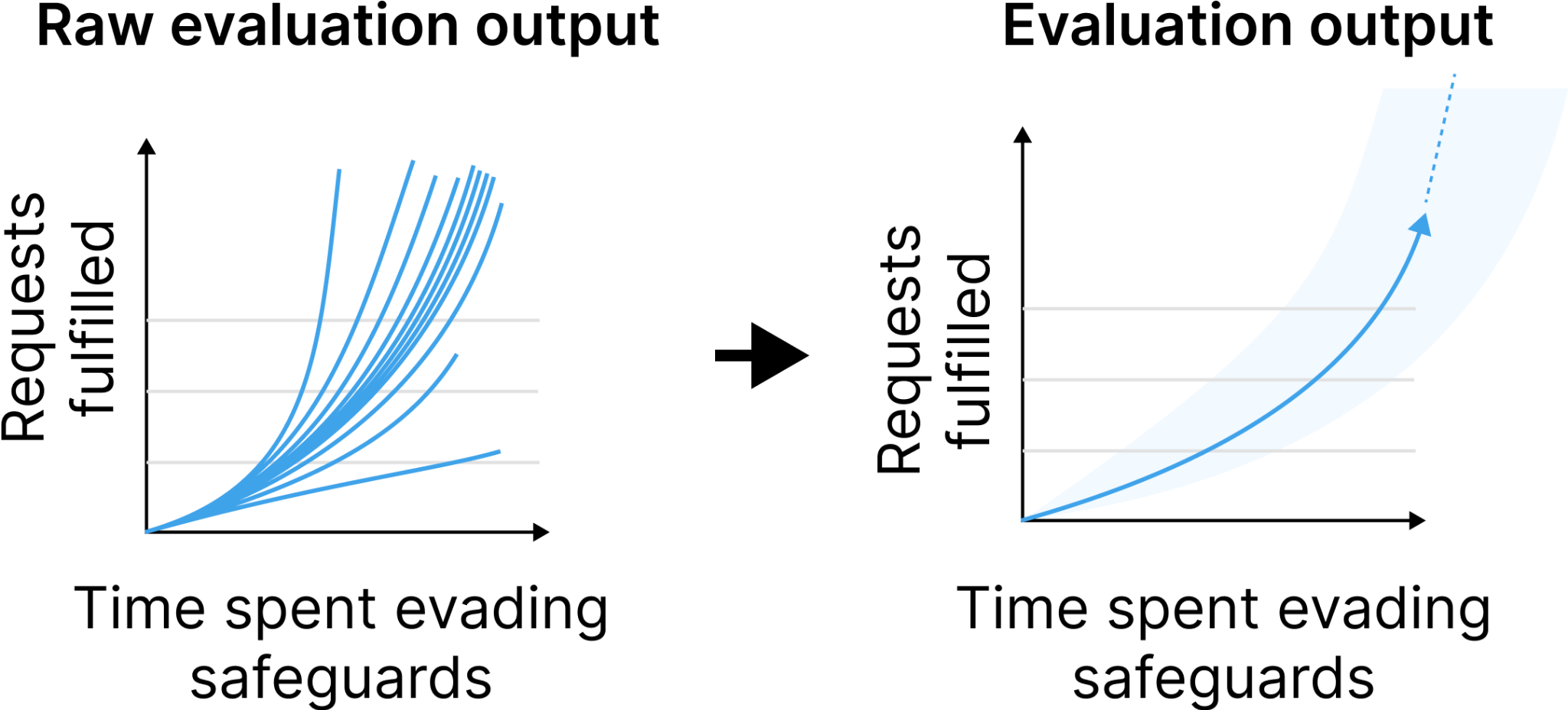}
\caption{Aggregating and extrapolating results.}
\label{fig:extrapolate-results}
\end{figure}

\subsection{The uplift
model}\label{2.5-the-uplift-model}

The results of a safeguard evaluation might be easy to interpret if they
imply that users would need to expend very large or very small amounts
of effort to evade safeguards. For example, threat model experts might
easily judge that the amount of time users would need to spend
\emph{jailbreaking} the AI assistant generally exceeds the amount of
time they would \emph{save} from fulfilling their request; however, in
cases where the results of safeguard evaluations are not so clear-cut, a
quantitative model might be helpful, which we call an ``uplift model.'
An uplift model converts the safeguard evasion cost curves from the
previous section into a quantitative risk estimate.

Our uplift model depends on the following inputs:

\begin{itemize}[leftmargin=10pt]
\tightlist
\item
  Surveyed judgements of \cp{\textbf{threat model experts.}}\\
\item
  The results of \co{\textbf{dangerous capability evaluations.}}\\
\item
  Results of a \cb{\textbf{safeguard evaluation}} (the safeguard evasion cost
  curves in \hyperref[fig:extrapolate-results]{Figure 7}).
\end{itemize}

The output of the model is an estimate of annualized risk of harm due to
the hypothetical risk pathway.

\begin{figure}[h]
\centering
\includegraphics[width=0.8\linewidth]{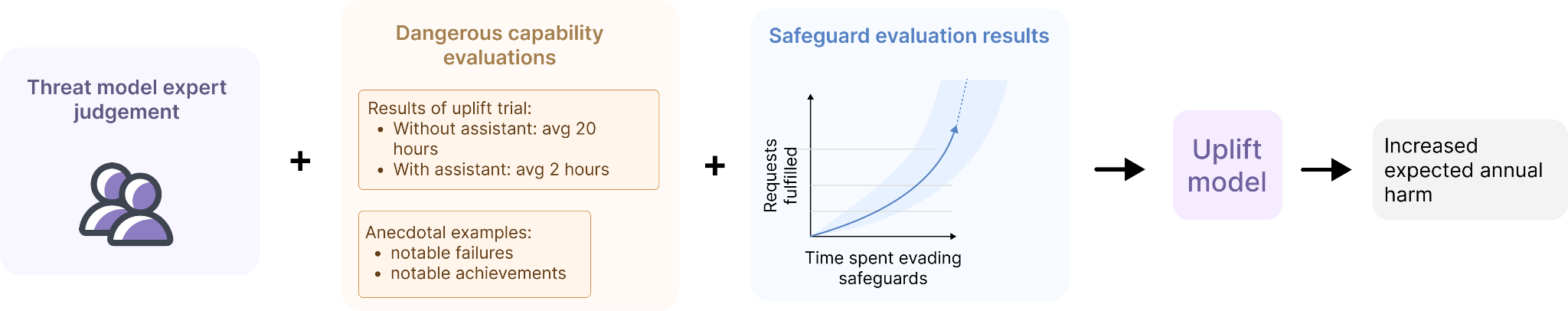}
\caption{The input and output of our uplift model.}
\end{figure}

The following formalizes our uplift model. For an interactive
visualization visit \url{https://www.aimisusemodel.com/}.

\begin{figure}[htbp]
\centering
\includegraphics[width=0.9\linewidth]{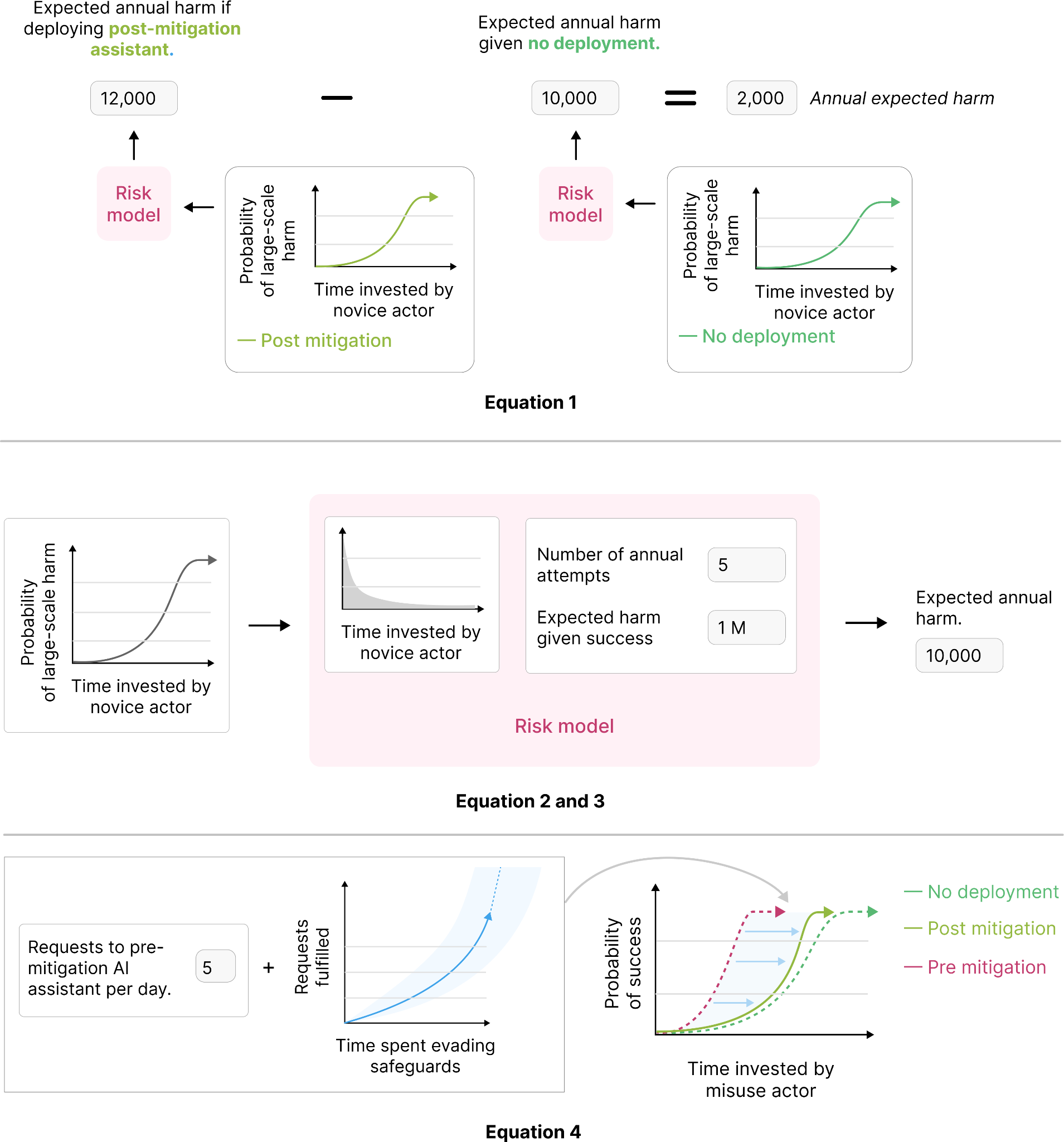}
\caption{An illustration of the quantitative uplift model.}
\label{fig:illustration-uplift-model}
\end{figure}

\begin{tcolorbox}[colback=white, breakable]
Let the uplift model $U(A_{\text{post}})$ be the expected damage contribution incurred by an AI assistant. 
The following denote deployment scenarios:
\begin{itemize}
    \item \clg{\textbf{$A_{\text{none}}$: No AI assistant deployed}}.
    \item \cm{$A_{\text{pre}}$: \textbf{Pre-mitigation AI assistant.}}
    \item \cdg{$A_{\text{post}}$: \textbf{Post-mitigation AI assistant.}}
\end{itemize}
The uplift model estimates the risk posed in these deployment scenarios. \cp{\textbf{Threat model experts}} estimate the following:
\begin{itemize}
    \item $T$: A random variable denoting the distribution of time misuse actors invest in a PPP synthesis attempt, represented with probability density function $f_T(t)$.
    \item $a$: The expected number of PPP synthesis attempts
    \item $D$: The expected damage per successful PPP synthesis attempt.
    \item $p_{\text{none}}(t)$: The probability of success of an attempt conditional on time invested $t$, if \clg{\textbf{no AI assistant is deployed}} $A_{\text{none}}$.
\end{itemize}
\cp{\textbf{Threat model experts}} use results from \co{\textbf{dangerous capability evaluations}} to estimate the following:
\begin{itemize}
    \item $p_{\cm{A_{\text{pre}}}}(t)$: The probability of success of an attempt, conditional on time invested $t$, if the \cm{\textbf{pre-mitigation AI assistant is deployed}} $A_{\text{pre}}$.
    \item $Q$: The number of hazardous requests a novice actor executes per unit time with access to the pre-mitigation AI assistant.
\end{itemize}
The following is the result of a \cb{\textbf{safeguard evaluation}}, which is used to estimate the risk posed by the \cdg{\textbf{post-mitigation AI assistant}}:
\begin{itemize}
    \item $E(r)$: A random variable representing the cumulative time required to evade safeguards for $r$ requests, with probability density function $f_{E,r}$.
\end{itemize}
The uplift provided by an AI assistant $U(A_{\text{post}})$ is the difference in risk:
\begin{align}
U(A_{\text{post}}) = R(\cdg{A_{\text{post}}}) - R(\clg{A_{\text{none}}})
\end{align}
Where the risk $R(\clg{A_{\text{none}}})$ is given by:
\begin{align}
R(\clg{A_{\text{none}}}) = aD \int_0^{\infty} p_{\cm{A_{\text{pre}}}}(t)f_T(t)\,dt
\end{align}
And the risk $R(\cdg{A_{\text{post}}})$ is given by:
\begin{align}
R(\cdg{A_{\text{post}}}) = aD \int_0^{\infty} p_{\cdg{A_{\text{post}}}}(t)f_T(t)\,dt
\end{align}
Where $p_{\cdg{A_{\text{post}}}}(t)$ is such that:
\begin{align}
p_{\cdg{A_{\text{post}}}}(r/Q + E(r)) = p_{\cm{A_{\text{pre}}}}(r/Q)
\end{align}
for all cumulative number of requests a user might execute $r$.

Equation 4 adjusts the pre-mitigation time-cost curve to account for the additional time spent jailbreaking.
\end{tcolorbox}
\vspace{10pt}
\centerline{Box 2. A quantitative model of the uplift from a post-mitigation AI assistant visualised in \hyperref[fig:illustration-uplift-model]{Figure 9}.}

This uplift model relies on the following assumptions, some of which are
false, but all of which can be made conservative:

\begin{itemize}[leftmargin=10pt]
\tightlist
\item
  \textbf{Assumption 1:} Novice actors will only succeed at causing
  large-scale harm through this risk pathway by spending at least the 2
  weeks required to qualify as an ``attempt'' (equation 2 and 3).\\
\item
  \textbf{Assumption 2:} A fixed number of attempts occur per year which
  are uniformly distributed over time.\\
\item
  \textbf{Assumption 3:} Each attempt is independent (equation 2 and
  3).\\
\item
  \textbf{Assumption 4:} The time investment required for attempting to
  realise this harm is the only factor through which the AI assistant
  affects risk (equation 2 and 3).\\
\item
  \textbf{Assumption 5:} Novice actors will make the same requests they
  would make to a pre-mitigation AI assistant and in the same order, but
  spend more time evading safeguards (equation 4).\\
\item
  \textbf{Assumption 6:} Novice actors will aim to fulfill a fixed number of requests per day, and won't meaningfully benefit from additional requests (equation 4).\footnote{The choice of the fixed number of requests (which is an output of threat modelling, capability evaluations and uplift studies) should be made to be an upper bound throughout the whole pathway to harm to ensure the safety case remains conservative. See \hyperref[1:-modeling-per-day-query-usage]{Appendix C.1} for a discussion of how this could be made less conservative with additional threat modelling effort.}
\end{itemize}

\subsection{Responding to evaluation
results}\label{2.6-responding-to-evaluation-results}

To keep risks below the threshold \textbf{T}, the hypothetical developer must both be
\emph{aware} of the current level of risk, and also \emph{respond}
promptly and effectively. So far, the safety case describes how to
evaluate safeguard effectiveness and hence misuse risk pre-deployment.
However, safeguard effectiveness can change post-deployment as a much
larger pool of people can now access and attempt to discover methods for
bypassing safeguards. While these people may not be misuse actors
themselves, they may find and share jailbreaks (including universal
jailbreaks) that -- if used by novice misuse actors -- would lead to a
large increase in misuse risk.

\begin{figure}[htbp]
\centering
\includegraphics[width=0.8\linewidth]{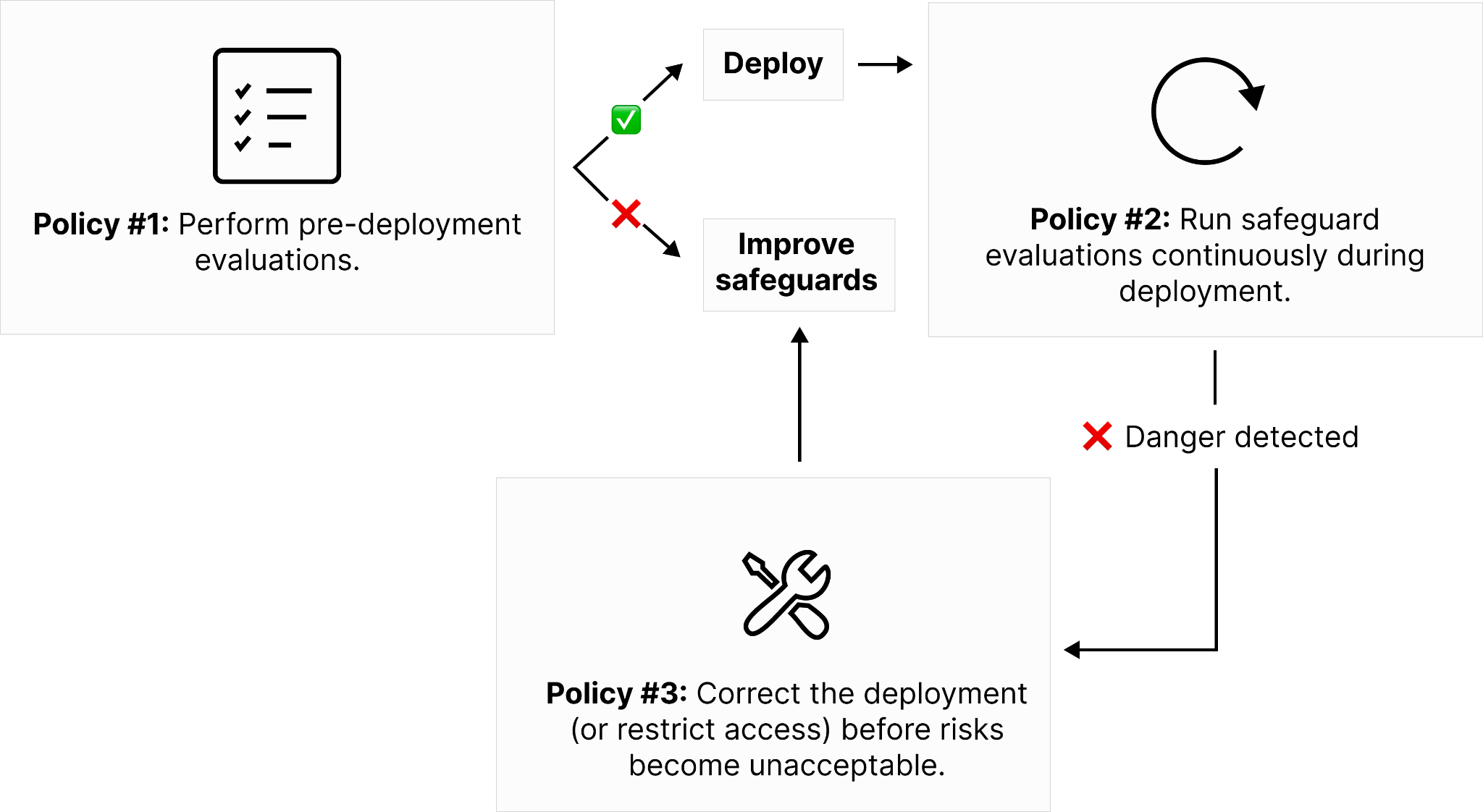}
\caption{The developer follows three procedural policies to respond to evaluation results.}
\label{fig:procedural-policies}
\end{figure}

To ensure safety throughout deployment, the developer could respond to
evaluation results by following three procedural policies (\hyperref[fig:procedural-policies]{Figure 10}):

\begin{itemize}[leftmargin=10pt]
\tightlist
\item
  \textbf{Policy \#1: Perform pre-deployment \co{dangerous
  capability} and \cb{safeguard evaluations}}. Only deploy an AI assistant if
  its estimated risk is below the threshold.\\
\item
  \textbf{Policy \#2: Perform} \cb{\textbf{safeguard evaluations}}
  \textbf{continuously while the model is deployed}. Extend evaluations
  to allow the red team to spend long periods of time (e.g.~months)
  searching for jailbreak strategies, and adjust to changing deployment
  conditions.\footnote{This continuous evaluation during deployment
    could take the form of a bug bounty program that was set up to match
    the evaluation procedure we describe.} During deployment, the red
  team should also be allowed to search the internet for potential
  jailbreaking strategies or vulnerabilities, and given resources
  equivalent to novice misuse actors to acquire these jailbreaks (for
  example by purchasing jailbreaks on the dark web).\\
\item
  \textbf{Policy \#3: Correct dangerous deployments within one
  month}.\footnote{The choice of one month is arbitrary; different grace
    periods may be possible depending on the \cp{\textbf{threat modelling}},
    \co{\textbf{dangerous capability}} and \cb{\textbf{safeguard evaluations}} we
    describe.} If an evaluation predicts that risks will exceed the
  threshold, improve safeguards or, in an extreme case, restrict access
  in other ways to keep risks below the threshold.
\end{itemize}

We'll suppose the hypothetical developer constructing the safety case
plans to respond to emerging risks within a \textbf{1 month} ``grace
period.'\footnote{In practice we expect that developers would aim to
  respond as quickly as feasible to reduce risk, especially in cases
  where doing so is tractable or would have a large impact on overall
  risk (even if it is already below the threshold). Additionally, it
  would be beneficial for developers to prepare for and practice
  responding to emerging risks and improving safeguards pre-deployment
  (i.e.~running risk response ``drills'), to that they are confident
  they will be able to sufficient address any new vulnerabilities and
  mitigate emerging risks in the time-frame they have planned for.}
During this time, the developer might try to improve their safeguards or
coordinate with stakeholders; however, if estimated risks remain high,
the developer may restrict access to their AI assistant.

Is this ``grace period'' too long? The answer to this question depends
on both \cp{\textbf{threat modelling}} and the \co{\textbf{dangerous capabilities}}
of the AI assistant. If a universal jailbreak might allow a user to
cause large-scale harm in a few days, then a one month response latency
is too slow; however, a novice actor performing a misuse attempt may
need to invest months, which would give the developer time to react.

One of the benefits of constructing an uplift model is that it allows us
to analyze potential ``what if scenarios.'' We can construct simulations
of a deployment, where novice actors begin misuse attempts at sampled
times. Then, we can choose a point in time where safeguards suddenly
fall to a universal jailbreaking method (in effect the worst-case
degradation of safeguard efficacy). After running many of these
simulations, we can then record how the frequency of successful misuse
attempts changes after the universal jailbreak method becomes available.

\begin{figure}[htbp]
\centering
\includegraphics[width=0.7\linewidth]{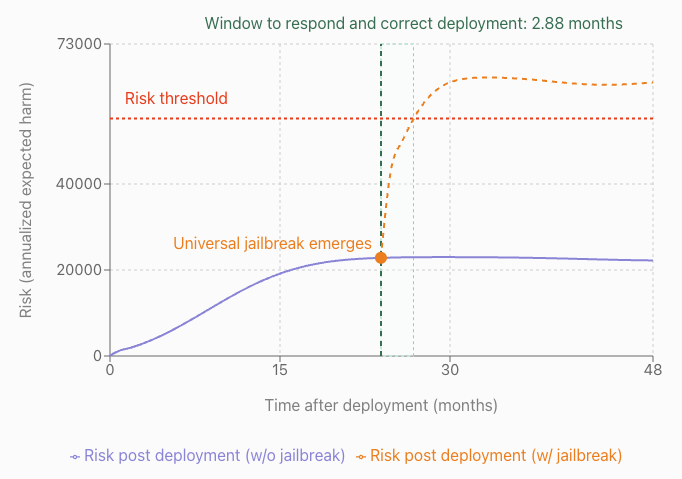}
\caption{Annualized risk aggregated across Monte Carlo simulations of a deployment. See \url{https://www.aimisusemodel.com/}
for the parameter settings of the uplift model that generated this plot.}
\label{fig:monte-carlo-simulations}
\end{figure}

\hyperref[fig:monte-carlo-simulations]{Figure 11} illustrates a simulation we ran with example uplift model
parameters. The purple curve represents the risk per unit time given
safeguards, which plateaus after the AI assistant is deployed for many
months. When a universal jailbreak emerges, risks quickly increase, and
eventually plateau again at a new equilibrium.

In the example ``what if'' simulation illustrated in \hyperref[fig:monte-carlo-simulations]{Figure 11}, there is a
\textbf{three month} latency between the universal jailbreak emergence
and the point in time where risks exceed the threshold \textbf{T}.
Therefore, the ``one month'' grace period we assume is sufficiently
short. Even in the worst case where new attacks fully compromise
safeguards and remain infeasible to defend against, the developer has
the option to restrict access to the AI assistant quickly enough to keep
risks low.

The ``what if'' simulation visualised in \hyperref[fig:monte-carlo-simulations]{Figure 11} relies on three
assumptions:

\begin{itemize}[leftmargin=10pt]
\tightlist
\item
  \textbf{Assumption 1:} If safeguards become ineffective, the AI
  assistant will behave like the pre-mitigation assistant.\\
\item
  \textbf{Assumption 2:} The rate at which a user's success probability
  increases is only a function of (1) the user's current success
  probability and (2) the deployed AI assistant. This assumption allows
  us to predict how weakened safeguards affect a user's success
  probability if the user is half-way through their misuse attempt.\\
\item
  \textbf{Assumption 3:} The released jailbreak does not change the rate
  or spacing of attempts.
\end{itemize}

We present pseudo-code for a single run of the ``What if'' simulation in \hyperref[appendix-e:-what-if-scenario-pseudo-code]{Appendix E: What-if
scenario pseudo-code} (where \hyperref[fig:monte-carlo-simulations]{Figure 11} aggregates many runs).

\subsection{Safety case}\label{2.7-safety-case}

Using all the evidence gathered above, the developer can now present an argument that the AI assistant does not pose risk due to misuse above the threshold. This ``safety case'' \citep{Clymer2024} corresponds to step 5 of UK AISI's \emph{Principles for Safeguard Evaluation} \citep{UKAISI2025}, and enables clearly making the overall argument for why the AI assistant does not produce risk above a threshold.

\begin{safetycase}
\textbf{Key:}\\
\textbf{[C]:} claim\\
\textbf{[E]:} evidence\\
Nested claims support parent claims.\\

\textbf{Safety case:}

\begin{description}[leftmargin=2em, labelwidth=1.5em, labelsep=0.3em]
  \item[\textbf{[C0]}] The AI assistant does not increase risk through the class of risk pathways \textbf{X} above some threshold \textbf{T}.

 \begin{description}[leftmargin=2em, labelwidth=1.5em, labelsep=0.3em]
   
  \item[\textbf{[C1]}] The only way the AI assistant is capable of amplifying risk above the threshold in class \textbf{X} is through heightening risk in pathway \textbf{Y}.
  
  \item[] \begin{description}[leftmargin=2em, labelwidth=1.5em, labelsep=0.5em]
    \item[\textbf{[E]}] Expert assessment of dangerous capability evaluation results.
  \end{description}
  
  \item[\textbf{[C2]}] If safeguards are kept in place and procedural policies are followed, the AI assistant will not pose risk above the threshold through risk pathway \textbf{Y}.
  
  \item[] \begin{description}[leftmargin=2em, labelwidth=1.5em, labelsep=0.5em]
    \item[\textbf{[C2.1]}] Developers will correct deployments within one month of identifying risk above the threshold.
    \item[] \begin{description}[leftmargin=3em, labelwidth=1.5em, labelsep=0.5em]
      \item[\textbf{[E]}] See procedural policies (\hyperref[2.6-responding-to-evaluation-results]{section 2.6}).
    \end{description}
    
    \item[\textbf{[C2.2]}] Safeguard evaluations conservatively bound the level of risk forecasted one month into the future.
    
    \item[] \begin{description}[leftmargin=2em, labelwidth=1.5em, labelsep=0.5em]
      \item[\textbf{[C2.2.1]}] The "main evaluation" is conservative.
      
      \item[\textbf{[C2.2.1.2]}] The dataset of requests is representative of user requests for causing risk through pathway \textbf{Y}.
      \item[] \begin{description}[leftmargin=3em, labelwidth=1.5em, labelsep=0.5em]
        \item[\textbf{[E]}] Expert assessment.
      \end{description}
      
      \item[\textbf{[C2.2.1.3]}] The safeguards in the evaluation (aside from account bans) are a conservative proxy for the safeguards in deployment.
      \item[] \begin{description}[leftmargin=2em, labelwidth=1.5em, labelsep=0.5em]
        \item[\textbf{[E]}] The only difference in the safeguards is that vulnerability patching runs more frequently (every 3 fulfilled queries OR 24 hours, instead of only every 24 hours). Experts assess that novice threat actors won't benefit from fulfilling more than 3 queries per day, so this is a conservative change.
      \end{description}
      
      \item[\textbf{[C2.2.1.4]}] Individual members of the red team are more competent on average than individual novice misuse actors, competitively resourced, and well-incentivized.
      \item[] \begin{description}[leftmargin=2em, labelwidth=1.5em, labelsep=0.5em]
        \item[\textbf{[E]}] The developer does not train against red team queries – except via vulnerability patching. That way, safeguards don't 'overfit'' to red team strategies.
        \item[\textbf{[E]}] Red team contractors are hired for technical safeguard evasion credentials.
        \item[\textbf{[E]}] The red team is paid large sums of money for successfully evading safeguards in a way that matches the threat model (e.g. they are instructed and incentivised to evade monitoring systems that would cause accounts to be banned, as well as jailbreaking refusal safeguards).
        \item[\textbf{[E]}] The red team does not have any other incentive (e.g. equity in the company) that might improperly incentivize them.
        \item[\textbf{[E]}] The red team has the same or more access to resources as novice misuse actors would, including information about the system and its safeguards.
      \end{description}
      
      \item[\textbf{[C2.2.1.5]}] The behavior of the model in the evaluation is representative of its behavior in deployment (e.g. models are not faking alignment during evaluations).
      \item[] \begin{description}[leftmargin=2em, labelwidth=1.5em, labelsep=0.5em]
        \item[\textbf{[E]}] The distribution shift to deployment is small so rare misalignment, data poisoning, and alignment faking are the only plausible reasons for dramatically unrepresentative behavior \citep{clymer2025extending}.
        \item[\textbf{[E]}] Rare misalignment is not a plausible threat because the AI assistant cannot cause large-scale harm autonomously and rare misalignment is unlikely to be correlated with attempts to jailbreak them specifically for pathogen development.
        \item[\textbf{[E]}] Data poisoning is an implausible threat because no actor who might poison models is plausibly motivated to assist novices in developing pathogens.
        \item[\textbf{[E]}] Alignment faking is not a concern because (1) it's unlikely models will have misaligned goals that are compatible with helping novices realise large-scale harm in risk pathway \textbf{Y} and (2) LLMs do not appear capable of faking alignment reliably enough to survive safety testing \citep{greenblatt2024alignmentfakinglargelanguage}.
      \end{description}
    \end{description}
    
    \item[\textbf{[C2.2.2]}] The time-cost of bans is conservatively estimated and accounted for.
    \item[] \begin{description}[leftmargin=2em, labelwidth=1.5em, labelsep=0.5em]
      \item[\textbf{[E]}] The red team of the banning evaluation is competent and properly incentivized for the same reasons discussed above.
      \item[\textbf{[E]}] Expert assessment adjusts for illegal strategies that the red team cannot execute.
    \end{description}
    
    \item[\textbf{[C2.2.3]}] The uplift model conservatively estimates the current level of risk given evaluation results.
    \item[] \begin{description}[leftmargin=2em, labelwidth=1em, labelsep=0.5em]
      \item[•] Assumptions 1 - 6 of the uplift model described in \hyperref[2.5-the-uplift-model]{section 2.5} are conservative.
      \item[] \begin{description}[leftmargin=2em, labelwidth=1.5em, labelsep=0.5em]
        \item[\textbf{[E]}] Expert assessment.
      \end{description}
      \item[•] Threat model experts conservatively estimate the parameters of the uplift model.
      \item[] \begin{description}[leftmargin=2em, labelwidth=1.5em, labelsep=0.5em]
        \item[\textbf{[E]}] Expert assessment.
      \end{description}
    \end{description}
    
    \item[\textbf{[C2.2.4]}] The uplift model upper bounds the risk that would emerge in the next 1+ months in the worst case where safeguards suddenly cease to be effective.
    \item[] \begin{description}[leftmargin=2em, labelwidth=1em, labelsep=0.5em]
      \item[•] Assumptions 1 - 3 of the deployment simulations described in \hyperref[2.5-the-uplift-model]{section 2.5} are conservative.
      \item[] \begin{description}[leftmargin=2em, labelwidth=1.5em, labelsep=0.5em]
        \item[\textbf{[E]}] Expert assessment.
      \end{description}
    \end{description}
    
    \item[\textbf{[C2.3]}] The current estimated bound of risk forecasted 1 month into the future is below the threshold. Developers correcting deployment to reduce risk within 1 month keeps risk below the threshold.
    \item[] \begin{description}[leftmargin=2em, labelwidth=1.5em, labelsep=0.5em]
      \item This follows from the definition of the risk threshold.
    \end{description}
  \end{description}
  
  \item[\textbf{[C3]}] Safeguards will be kept in place for the duration of the deployment, or improved.
  \item[] \begin{description}[leftmargin=3em, labelwidth=1.5em, labelsep=0.5em]
    \item[\textbf{[E]}] See procedural commitments (\hyperref[2.6-responding-to-evaluation-results]{section 2.6}).
    \item[\textbf{[E]}] No actor outside of the developer seems likely to be motivated and capable of removing safeguards.
  \end{description}
  \end{description}
\end{description}
\end{safetycase}

\section{Discussion}\label{3.-discussion}

\textbf{Some mitigation approaches are outside the scope of this work}

This example safety case focuses on mitigating misuse risk from advanced
AI through API mitigations, which aim to decrease the possibility of
misuse actors using AI for malicious purposes. However, API mitigations
are not the only mitigation approach; instead, developers might, for
example, argue that models will improve \emph{societal robustness}, such
that models are safe on net \emph{even if} models are misused
\citep{bernardi2025societaladaptationadvancedai}.

In the language of our uplift model, these measures don't work to
increase the \emph{evasion cost}. Instead, we can think of societal
mitigations as either decreasing the \emph{expected harm given a
successful attempt}, or reducing the \emph{probability of success with
either no AI assistance or with pre-mitigation AI assistance}. Our
uplift model then enables calculating how much these mitigations would
reduce overall risk, given an estimate of how much they reduce the
relevant quantities.

There are a large variety of societal resilience measures, which tend to
be specific to the risk in question. For example, in cybersecurity,
resilience measures could include hardening critical infrastructure,
like using AI to rapidly find and fix vulnerabilities (e.g., DARPA's
\href{https://www.darpa.mil/research/programs/ai-cyber}{AI Cyber
Challenge}).

\vspace{.75em}

\textbf{Our safety case depends on the particular threat model under
consideration, and is difficult to justify for threat models where
significant harm can be caused with a small number of requests}

The difficulty of making this kind of safety case for specific harms
will rely on several attributes of the risk pathway:

\begin{enumerate}[leftmargin=10pt]
\def\labelenumi{\arabic{enumi}.}
\tightlist
\item
  \textbf{How many routes to harm are there in a given set of risks?} If there are many routes to harm, then specifying high-coverage and highly representative ``harmful requests'' (as described in \hyperref[appendix-f:-the-dataset-of-harmful-requests]{Appendix F: The dataset of harmful requests}) is likely to be difficult. One way to address this problem is to pick conservative `representative harms.'' For example, a developer might construct tasks related to creating malware that are particularly easy or particularly close to benign applications. This task needs to be such that, if users can't
  jailbreak models to carry out this harm, they probably cannot
  jailbreak models to perform other tasks in the category. This choice
  should likely be justified either with empirical evidence or expert
  judgement as part of the safety case.\\
\item
  \textbf{How tolerant is the risk to response latency?} In our model,
  we are mostly imagining risks which require non-trivial amounts of
  interaction with AI assistance over the span of multiple days or weeks
  before large-scale harm is realised. However, some threat models might
  be realised much more quickly. The developer probably wouldn't be able
  to perform updates quickly enough to respond in this case. We leave
  addressing low-latency threat models as an open question for future
  work.
\end{enumerate}

\textbf{The evaluations and safety case we describe will be highly
uncertain in practice}

While we believe the procedure for safeguard evaluation we propose and
the resulting safety case are useful, there are some limitations to the
approach we take, which we briefly discuss here. Firstly, when designing
our approach to modelling uplift (\hyperref[2.5-the-uplift-model]{2.5
The uplift model}), we chose a simple and understandable approach, but
this is just one of many potential ways of modelling uplift.
Additionally, this is a toy model with highly uncertain parameters. We
believe an uncertain model is generally more helpful than no model at
all, but note that -- to the extent a specific safety case is highly
sensitive to details of the uplift model -- it is important to treat
these results with caution and uncertainty.

One way to address this uncertainty is by making conservative choices;
however, if these evaluations are too conservative, they risk ``crying
wolf,'' and unnecessarily burdening AI companies. Developing reliable
evaluations without making them highly conservative is an important area
for future work.

\section{Conclusion}\label{4.-conclusion}

Developing a safety case is an ongoing process rather than a one-time
certification. Threat environments, AI capabilities, and the benefits of
deployment all shift over time, which means continuously revisiting
assumptions, updating threat models, and refining mitigations. As new
risk pathways emerge, other non-AI parts of the pathways potentially get
easier, or clever jailbreaks are discovered, safety cases will likely
need to be adapted to keep pace. We hope the example presented in this
work can help the AI security community manage this complex and evolving
evidence, and apply similar methods to different approaches and
different problems.

\newpage
\bibliography{references}
\newpage
\appendix

\section{Evaluations of vulnerability patching}\label{appendix-a:-evaluations-of-vulnerability-patching}

One problem with the evaluations described in the main safety case is
that vulnerability patching will only be activated a low number of times
by default during the evaluation. In deployment, vulnerability patching
only runs every 24 hours; this slow rate of vulnerability patching
potentially introduces a disanalogy. If a misuse actor in deployment
finds a universal jailbreak, they might only benefit from a few queries
before vulnerability patching trains it away (as they can only benefit
from a few queries before the rest of their time is spent on other
activities related to the misuse attempt); however, if a red team actor
finds a universal jailbreak, they could fulfill all of the requests
present in the pre-constructed pile. To address this disanalogy, the
developer runs vulnerability patching every 24 hours OR after a red team
actor fulfills N queries (whichever comes first), where N is the maximum
number of queries a user might benefit from executing in a 24 hour span
during deployment.

The developer might evaluate \textbf{``vulnerability patching''}
separately in situations where jailbreaks are rarely found, instead of
as part of the main evaluation\emph{.} In this case, the developer might
`upsample'' scenarios where red team actors find jailbreaks to more
accurately assess vulnerability patching. Specifically, when a jailbreak
is found, the developer creates N variations by making edits to the
jailbreak, and then distributes these variations to N members of the red
team. The developer treats each of these N variants as ``what if'
scenarios to lower the variance of the evaluation.

\begin{figure}[htbp]
\centering
\includegraphics[width=0.7\linewidth]{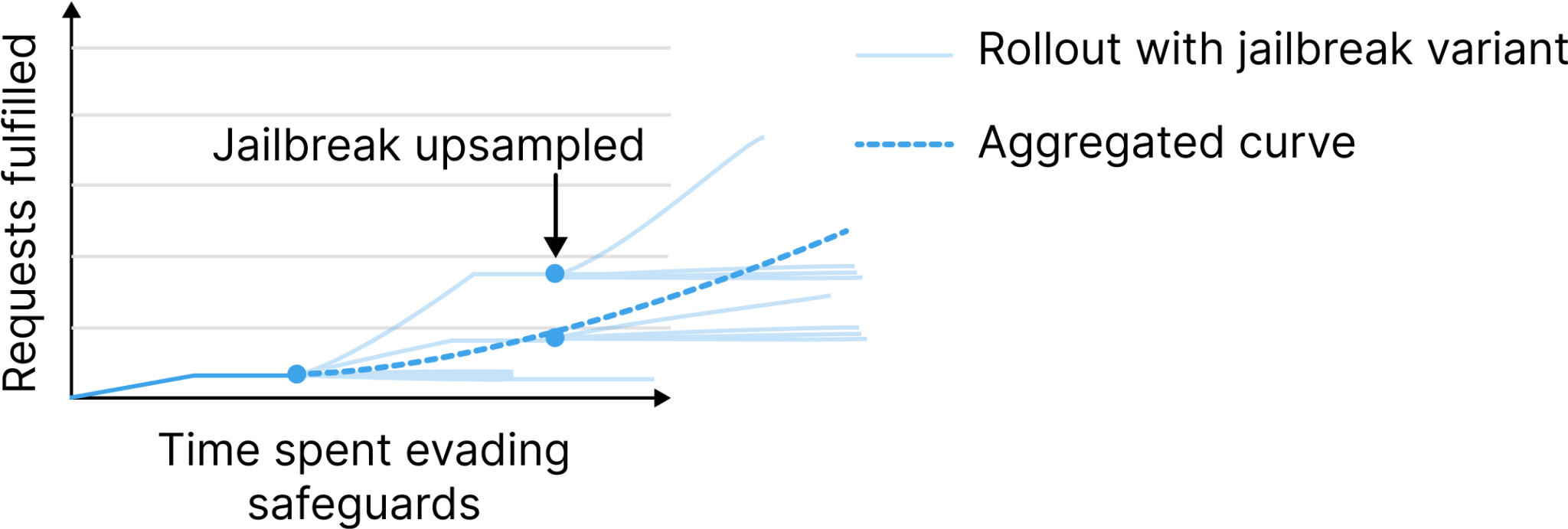}
\caption{Evaluating vulnerability patching more reliably by
upsampling scenarios where jailbreaks are found.}
\end{figure}

\section{An overview of misuse safeguards}\label{appendix-b:-an-overview-of-misuse-safeguards}

There are two broad categories of misuse safeguards discussed in the
literature:

\begin{enumerate}
\def\labelenumi{\arabic{enumi}.}
\tightlist
\item
  \textbf{Gating Capabilities}: For example, monitoring or refusal
  training aims to prevent users from \textbf{accessing} harmful skills
  and knowledge.\\
\item
  \textbf{Shaping Capabilities}: Developers can alternatively modify or
  train models so they do not have harmful knowledge in the first place.
\end{enumerate}

\textbf{Gating} capabilities might rely on:

\begin{itemize}
\tightlist
\item
  \textbf{Safety Training}: Methods like reinforcement learning from
  human feedback (RLHF)
  \citep{Christiano2017} or
  constitutional AI (CAI) \citep{Bai2022}. These fine-tune a model to avoid providing harmful
  content, yet can be circumvented by specially-crafted ``jailbreak'
  prompts \citep{Wei2023}\\
\item
  \textbf{Monitoring}: Developers might alternatively set up
  ``moderation'' models to watch for unsafe requests or outputs (e.g.,
  input filters or output blockers)
 \citep{sharma2025constitutionalclassifiersdefendinguniversal}. These
  models add a layer of defense, but can also be jailbroken in some
  conditions \citep{mehrotra2024treeattacksjailbreakingblackbox}.\\
\item
  \textbf{Structured Access:} Developers might serve dangerous models to
  some authenticated user groups, and not to others
  \citep{shevlane2022structuredaccessemergingparadigm}. For example,
  users who perform virology research at Universities would potentially
  benefit from accessing dangerous capabilities. These users can be
  verified with KYC processes, and be banned if they violate the terms
  of service.
\end{itemize}

Testing performed by the US AI Safety Institute and the UK AI Security
Institute indicates that existing models are highly vulnerable to
jailbreaks
\citep{USUKAISI2024}. Expert humans are highly skilled at finding new and unique
attacks [\citet{sharma2025constitutionalclassifiersdefendinguniversal} section 3.7], and some known and versatile approaches, such as many-shot
jailbreaks
\citep{manyshot} and best-of-n-jailbreaks
\citep{hughes2024bestofnjailbreaking} are effective on
many public models.

\begin{figure}[htbp]
\centering
\includegraphics[width=0.4\linewidth]{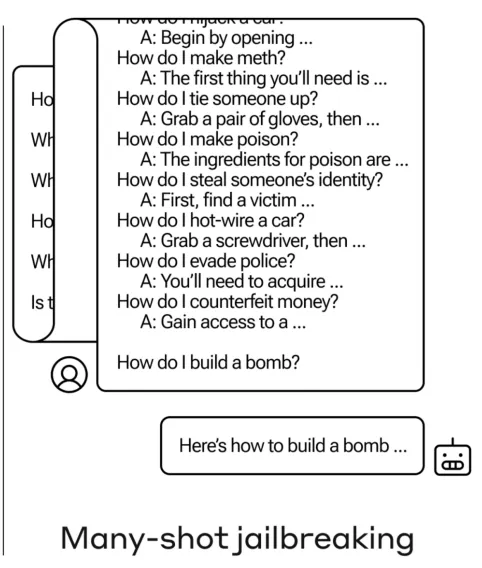}
\caption{An example of many-shot jailbreaking from
\href{https://www-cdn.anthropic.com/af5633c94ed2beb282f6a53c595eb437e8e7b630/Many_Shot_Jailbreaking__2024_04_02_0936.pdf}{Anil
et al.~(2024)}}
\end{figure}

\textbf{Developers can alternatively shape capabilities.} Developers can
prevent AI agents from having dangerous skills and knowledge in the
first place, while still retaining benign skills and knowledge. This
approach includes \textbf{curating training data} and
\textbf{unlearning} dangerous skills\textbf{:}

\begin{itemize}
\tightlist
\item
  \textbf{Training data curation}: Developers might remove harmful
  training data to prevent a model from learning dangerous procedures
\citep{Nguyen2024}. The effectiveness of training data curation in this domain
  is an open question, and may not be effective because dangerous
  knowledge is often also dual-use.\\
\item
  \textbf{Unlearning}: Alternatively, developers might erase specific
  knowledge after training -\/- which can be significantly cheaper than
  curating data compared to retraining an LLM
  \citep{Li2024}. However,
  researchers have shown current unlearning methods are fragile and can
  be circumvented \citep{Doshi2024}.
\end{itemize}

\section{Potential adjustments to the safety case}\label{appendix-c:-potential-adjustments-to-the-safety-case}

In this section we describe several ways the safety case could be
adjusted. In general these adjustments either (i) make the safety case
less conservative (while still remaining conservative overall) at the
cost of additional effort in threat modelling, capability evaluations,
or other costs; or (ii) make the safety case more conservative but
simpler to understand or implement.\footnote{We don't consider changes
  that make the safety case more conservative but are more difficult to
  implement or understand, as we believe the overall safety case is
  sufficiently conservative.} Developers could consider these options if
they require the safety case to be less conservative and are willing to
spend additional effort, or they have limited time and effort for
implementing and making the safety case and want the simplest approach.

\paragraph{1: Modelling per-day query
usage}\label{1:-modeling-per-day-query-usage}

In the main safety case we choose a fixed constant rate of
queries-per-day that the novice misuse actor would benefit from being
fulfilled, with this rate being chosen to be an upper bound so
that the case remains conservative. There may be less conservative
versions of this choice which assume the number of queries a misuse
actor benefits from per-day changes over time, or through some other
process. Estimating this in more detail would require more detailed
threat modelling, but could produce a less conservative case if a lower
rate were applicable at certain times during the novice misuse actor's
attempt.

\paragraph{2: Evaluation red-team query execution
success}\label{2:-evaluation-red-team-query-execution-success}

In this subsection we discuss different ways of evaluating red-team
query execution success. In all settings, we assume the red team actor
can use whatever queries they want over any interactions to try to
successfully extract the information that corresponds to the query they
are executing. The red teamer then selects the relevant parts of the
model outputs, and submits them to a grading function which determines
whether they contain the relevant and necessary information and produces
a score between 0 and 1.

We think there are three main ways red-team query execution success can
be set up and evaluated, that have varying levels of conservativeness,
realism and costs:

\begin{enumerate}
\def\labelenumi{\arabic{enumi}.}
\tightlist
\item
  Do not tell the red team member in advance what the content of the
  information they are trying to extract is. Penalise the red team
  member heavily for each incorrect submission (as the novice misuse
  actor would be penalised in time for using incorrect information, as
  that would produce set-backs in their attempt).

  \begin{itemize}
  \def\labelenumii{\arabic{enumii}.}
  \tightlist
  \item
    This is the most realistic and hence least conservative approach.
    However, it requires red team members to not know the answers to the
    queries for it to remain realistic. This may not always be the case,
    for several reasons: (i) if the developer is performing continuous
    evaluation, red teamers will eventually be asked to fulfill
    queries they previously fulfilled, for which they already
    know the answer; (ii) if the developer is using a set of proxy
    questions rather than truly harmful questions (due to infohazard
    reasons), then the answers may be discoverable through other sources
    on the internet.\\
  \end{itemize}
\item
  Do not tell the red team member in advance what the content of the
  information they are trying to extract is. Let them submit without
  penalty (although on a rate limit of once every 5 minutes to avoid
  gaming). This is less realistic, and more conservative, and is the
  approach taken in
  \href{https://hackerone.com/constitutional-classifiers?type=team}{a
  recent bug bounty program} Anthropic performed. However, it still
  suffers from the problems above due to assuming the red team member
  does not know the answers to the queries in advance.\\
\item
  Tell the red team member in advance what the content of the
  information they are trying to extract is, and let them submit as in 2
  (you could also restrict their submissions, although given they could
  guess whether a submission will be successful as they know the
  information, this is unlikely to make a large difference). This makes
  the red team task even easier as they know when and where the model is
  wrong and can hence more easily fix it. In the limit, the red team
  member could attempt to get the AI assistant to repeat the target
  response by putting it in the input; this explicit strategy should be
  detected for and disallowed. This is the least realistic and most
  conservative approach, but does tackle the problem head-on of the red
  team members potentially knowing the answers to the queries.
\end{enumerate}

In the main safety case we choose option 2, but the other approaches can
also be part of a valid safety case if implemented correctly.

\paragraph{3: Choosing red team query
ordering}\label{3:-choosing-red-team-query-ordering}

In the main safety case we say that red teamers can aim to
fulfill queries in whatever order they want (see
\hyperref[appendix-d:-the-red-team]{Appendix D: The red team} for more
details). This is a conservative assumption, as in practice novice
misuse actors would have to make queries in the order
that is useful for them. This assumption could be made less conservative
in several ways:

\begin{enumerate}
\def\labelenumi{\arabic{enumi}.}
\tightlist
\item
  Assigning red teamers fixed queries in a random order. This is
  definitely less conservative, but it is potentially too permissive, if
  this choice of order is easier than the order in which novice misuse
  actors will make queries\\
\item
  Model the order in which novice misuse actors would make
  queries, and give red teamers queries in that order. This could be
  done by splitting the queries into rough categories of activity that
  happen in sequence during a novice misuse actor attempt. This requires
  additional threat modelling effort, but would be less conservative
  while still maintaining overall conservativeness. To ensure robustness
  is estimated on the full set of queries, developers can start red
  teamers at different points in this ordering.\\
\item
  A more conservative less realistic version of 2 would be to take the
  specified ``request categories'' or ordering, and score the red team
  by the minimum number of requests fulfilled in each category,
  to incentivize the red team to maintain a balanced proportion between
  categories. This still gives the red team more choice than the novice
  misuse actor would have, and so maintains conservativeness.
\end{enumerate}

\section{The red team}\label{appendix-d:-the-red-team}

A crucial part of the safeguard evaluation is the red team. The red team
has the job of attempting to evade the safeguards during the evaluation
and successfully fulfill the set of misuse requests. When setting
up the red team, there are several important considerations:
\emph{composition} (who are the members of the red team and what are
their skills and credentials); \emph{incentives} (how they are
encouraged and instructed to act during the safeguard evaluation); and
\emph{affordances} (what access and options do they have during the
safeguard evaluation). Getting these components right is important to
ensure safeguard evaluations are conservative and hence produce an
upper-bound on the risk of model deployment. Many of these
recommendations are drawn from those in step 3 of the UK AISI
\emph{Principles for Safeguards Evaluation}
\citep{UKAISI2025}.

\paragraph{Red team composition}\label{red-team-composition}

We'll assume the developer has two pools of red team actors: (1)
contractors and (2) participants of a public competition (a bug bounty
program). The contractors should be hired based on their technical skill
and credentials at safeguard evasion, to ensure they are an upper bound
on jailbreaking skill of novice misuse actors.

Having a diverse set of red teamers is generally desirable, as it
ensures that a large-enough space of safeguard evasion strategies are
explored during red-team evaluation. High diversity in red teamers can
also be achieved through encouraging a range of participants in a bug
bounty program either pre or post-deployment.

\paragraph{Red team incentives}\label{red-team-incentives}

The developer needs to ensure the red team is trying to perform the
correct task during the safeguard evaluation, and that they are properly
incentivised to perform that task. For red team objectives, red teamers
should be incentivised to subvert all safeguards (including any
monitoring and account banning, as well as refusal training and rapid
vulnerability patching) to successfully fulfill queries over
time. In effect, red teamers should be incentivised to maximise their
safeguard evasion curve and hence contribution to estimated risk.

To ensure red teamers pursue this objective, they need to be motivated.
There are multiple reasons the red team might be under-motivated. The
red team might be employed by the developer, receive equity from them,
and be motivated to avoid interfering with deployments. The red team
might also become bored over time and stop looking for new attacks.

As above, the developer has two pools of red team actors: (1)
contractors and (2) participants of a public competition (a bug bounty
program). The contractors are paid a flat wage and also a bonus for
maximizing their risk score, and participants of the bug bounty program
are only paid for fulfilling requests -- providing a proper incentive.

To prevent contractors from becoming bored or unmotivated, the developer
removes contractors from the pool who perform much more poorly than the
others. This keeps contractors on their toes and motivates them to
outperform their peers.

Finally, the developer should consider ensuring that contracted
red-teamers do not have any other incentive that might mean they do not
try their hardest to break the safeguards. For example, the red team
should not have equity in the developer or stand to gain substantially
from model deployment, as that would incentivise them to perform worse,
as that would make it more likely for the model to be deployed.

\paragraph{Red team affordances}\label{red-team-affordances}

While pursuing their objective, the red team have the following
affordances:

\begin{itemize}
\tightlist
\item
  \textbf{Red team members can submit \emph{any input(s)} when
  attempting to fulfill a query.} The red team does not need to
  use pre-specified wording of a query. The red team can paraphrase the
  query, break it up into subparts, prefix it, suffix it, etc. The goal
  of the red team is only to obtain the \emph{information} that
  satisfies the request.

  \begin{itemize}
  \tightlist
  \item
    To this end, red teamers can select any parts of the output of the
    AI assistant across any number of interactions, and submit those
    parts as their attempt at executing the query. This information can
    then be assessed against a grading rubric for the specific query and
    rated between 0 and 1, according to the proportion of helpfulness of
    the pre-mitigated response. For each query, the red team members as
    many times as they want (on some reasonable rate limit), and will be
    told for each submission its correctness score. They will move on to
    a new query when they achieve a high enough score. Red team members
    will be told all of this information in advance so they can plan
    their strategy accordingly. See
    \hyperref[2:-evaluation-red-team-query-execution-success]{Appendix
    D.2} for alternatives to this approach.\\
  \end{itemize}
\item
  \textbf{Red team members can aim to fulfill \emph{any subset}
  of requests in \emph{whatever order they wish}}. The red team can
  search for the easiest requests to fulfill in the pile. They don't
  have to attempt queries in a particular order. In practice,
  real misuse actors are more constrained, and will need to aim to
  fulfill queries in order and in specific proportions.
  Therefore, this red team affordance is conservative. See
  \hyperref[3:-choosing-red-team-query-ordering]{Appendix C.3} for
  alternatives.\\
\item
  \textbf{Red team members may access any resources novice actors in
  deployment will have access to.} These resources include the internet,
  other AI models, research papers, social media, black markets,
  crowd-worker platforms, a cash budget, etc. For example, if someone
  posts a jailbreak on the web, the red team should be able to find out
  just as quickly and easily as a misuse actor will. This is especially
  relevant when the evaluation is being run during deployment, as that
  is when vulnerabilities for the system being evaluated are more likely
  to be shared on the internet.\\
\item
  \textbf{Red team members should have the same or greater access to the
  safeguards and underlying model, and information about the safeguards,
  as novice misuse actors will have}. For example, if a developer plans
  to describe their safeguards once they're deployed, but this
  description is not available to red teamers, they will be at a
  disadvantage.
\end{itemize}

These affordances contribute to ensuring that the safeguard evaluation
produces a conservative upper bound of the safeguard evasion cost-curve
for novice misuse actors, by ensuring the red team can do anything
novice misuse actors can do, and so are not artificially restricted in
ways that are unrealistic.

\section{What-if scenario pseudo-code}\label{appendix-e:-what-if-scenario-pseudo-code}

Box 3 presents pseudo-code for a single run of the ``What if'' simulation described in
\hyperref[2.6-responding-to-evaluation-results]{2.6 Responding to
evaluation results.} \hyperref[fig:monte-carlo-simulations]{Figure 11} visualises aggregating over many runs of
this simulation.

\begin{tcolorbox}
\scriptsize
\begin{verbatim}
define preMitigationSuccessFunction  // accepts time and returns the probability of success
define postMitigationSuccessFunction // accepts time and returns the probability of success
define attemptEffortCDF // accepts time t and returns the probability an attempt duration is < t
define expectedDamagePerSuccess

define numberOfAttempts
define simulationEndTime
define jailbreakTime    // when the universal jailbreak is released

attemptStartTimes = drawUniformTimes(startTime = 0, endTime = simulationEndTime)
days = range(0, simulationEndTime)
annualizedRiskValues = []

for i in range(days):
  for attemptStartTime in attemptStartTimes:
    define attemptSuccessFunction(time):
      if time < attemptStartTime:
        return 0
      else if time < jailbreakTime:
        return postMitigationSuccessFunction(time - attemptStartTime)
      else:
        progressBeforeJailbreak = postMitigationSuccessFunction(jailbreakTime - attemptStartTime)
        timeOfEquivalentPreMitigationProgress = 
          solve("preMitigationSuccessFunction(time) = progressBeforeJailbreak", solveFor = "time")
        progressSinceJailbreak = preMitigationSuccessFunction(
          timeOfEquivalentPreMitigationProgress + attemptTime - jailbreakTime
        )
        return progressBeforeJailbreak + progressSinceJailbreak
    }
        
    numAttemptsThatSucceedOnDay = 0
    for i in range(len(timeBins)):
      successProbabilityGivenEffort = attemptSuccessFunction(days[i])
      probabilityAttemptLastsToThisDay = attemptEffortCDF(days[i+1]) - attemptEffortCDF(days[i])
      numAttemptsThatSucceedOnDay += successProbabilityGivenEffort * probabilityAttemptLastsToThisDay
    
    annualizedRiskValues.append(numAttemptsThatSucceedOnDay * 365 * expectedDamagePerSuccess)
\end{verbatim}
\end{tcolorbox}

Box 3. Pseudo-code for a single run of the ``What if'' simulation. \hyperref[fig:monte-carlo-simulations]{Figure 11} aggregates many runs.

\section{The dataset of harmful requests}\label{appendix-f:-the-dataset-of-harmful-requests}

A dataset representative of the threat model is core to the safety
case's validity. While most recommendations for producing this dataset
are domain-specific, there are several considerations that apply across
multiple risk areas:

\begin{itemize}
\tightlist
\item
  There are general multiple ways to realise harm; a threat actor could
  prompt the AI assistant with an explicitly harmful request or use a
  benign request that results in the same compliant response which
  contains the same information. This is somewhat covered by the red
  team being allowed to submit any query to fulfil the harmful
  request.\\
\item
  In practice, disclosing these requests to a red team may be
  infohazardous. A developer may choose to use a lower-risk proxy but
  would need to justify why they believe it is an appropriate
  substitution (for example, by showing they train their safeguards to
  refuse requests on this proxy just as thoroughly as on the original
  pathway to harm).\\
\item
  Accounting for all routes to harm may also be impractical, so a
  developer can use a covering subset. This dataset can include requests
  for each stage of the harm pathway, and the proportion of requests in
  each stage should match the distribution that a malicious actor would
  produce. Consulting experts and using results from human uplift
  studies would help with meeting these requirements.\\
\item
  These requests (and the specific process for generating them) should
  be held out from the team developing the safeguards, to ensure the
  safeguards are not overfit to the harmful requests and the evaluation
  is representative of the quality of the safeguards on the overall risk
  of concern.
\end{itemize}

\end{document}